\documentclass{article}

\usepackage{arxiv}

\usepackage[utf8]{inputenc} 
\usepackage[T1]{fontenc}    
\usepackage[hidelinks]{hyperref}
\usepackage{url}            
\usepackage{booktabs}       
\usepackage{amsmath}
\usepackage{amsfonts}       
\usepackage{nicefrac}       
\usepackage{microtype}      
\usepackage{cleveref}       
\usepackage{lipsum}         
\usepackage{graphicx}
\usepackage{caption}
\usepackage{subcaption}
\usepackage{natbib}
\usepackage{doi}

\title{M2N: Mesh Movement Networks for PDE Solvers}

\author{
	Wenbin Song \\
	ShanghaiTech University \\
	Shanghai 201210, China \\
	\texttt{songwb@shanghaitech.edu.cn} \\
\And
	Mingrui Zhang \\
	Imperial College London \\
	London, SW7 2AZ, UK \\
	\texttt{mingrui.zhang18@imperial.ac.uk} \\
\And
	Joseph G. Wallwork \\
	Imperial College London \\
	London, SW7 2AZ, UK \\
	\texttt{j.wallwork16@imperial.ac.uk} \\
\And
	Junpeng Gao \\
	ETH Zürich \\
	8092 Zurich, Switzerland \\
	\texttt{jungao@student.ethz.ch} \\
\And
	Zheng Tian \\
	ShanghaiTech University \\
	Shanghai 201210, China \\
	\texttt{zheng.tian.11@ucl.ac.uk} \\
\And
	Fanglei Sun \\
	ShanghaiTech University \\
	Shanghai 201210, China \\
	\texttt{sunfl@shanghaitech.edu.cn} \\
\And
	Matthew D. Piggott \\
	Imperial College London \\
	London, SW7 2AZ, UK \\
	\texttt{m.d.piggott@imperial.ac.uk} \\
\And
	Junqing Chen \\
	Tsinghua University \\
	Beijing 100084, China\\
	\texttt{jqchen@tsinghua.edu.cn} \\
\And
	Zuoqiang Shi \\
	Tsinghua University \\
	Beijing 100084, China\\
	\texttt{zqshi@tsinghua.edu.cn} \\
\And
	Xiang Chen \thanks{Corresponding Authors: Xiang Chen and Jun Wang.}\\
	Noah’s Ark Lab, Huawei \\
	Beijing 100084, China\\
	\texttt{xiangchen.ai@outlook.com} \\
\And
	Jun Wang \footnotemark[1]\\
	University College London \\
	London WC1E 6BT, UK \\
	\texttt{jun.wang@cs.ucl.ac.uk}
}

\renewcommand{\shorttitle}{M2N: Mesh Movement Networks for PDE Solvers}

\hypersetup{
pdftitle={M2N: Mesh Movement Networks for PDE Solvers},
pdfsubject={q-bio.NC, q-bio.QM},
pdfkeywords={First keyword, Second keyword, More},
}

\begin{document}
\maketitle

\begin{abstract}
    Mainstream numerical Partial Differential Equation (PDE) solvers require discretizing the physical domain using a mesh. Mesh movement methods aim to improve the accuracy of the numerical solution by increasing mesh resolution where the solution is not well-resolved, whilst reducing unnecessary resolution elsewhere. However, mesh movement methods, such as the Monge-Amp\`ere method, require the solution of auxiliary equations, which can be extremely expensive especially when the mesh is adapted frequently. In this paper, we propose to our best knowledge the first learning-based end-to-end mesh movement framework for PDE solvers. Key requirements of learning-based mesh movement methods are alleviating mesh tangling, boundary consistency, and generalization to mesh with different resolutions. To achieve these goals, we introduce the neural spline model and the graph attention network (GAT) into our models respectively. While the Neural-Spline based model provides more flexibility for large deformation, the GAT based model can handle domains with more complicated shapes and is better at performing delicate local deformation. We validate our methods on stationary and time-dependent, linear and non-linear equations, as well as regularly and irregularly shaped domains. Compared to the traditional Monge–Amp\`ere method, our approach can greatly accelerate the mesh adaptation process, whilst achieving comparable numerical error reduction.
\end{abstract}


\section{Introduction}
\label{section:intro}


Partial Differential Equations (PDEs) are widely used to model natural phenomena, ranging from aerospace engineering, atmosphere/ocean dynamics, bio-engineering to computer graphics. Efficient and accurate numerical solutions for complex PDEs are essential but challenging problems in all scientific and engineering disciplines. 

Solving PDEs using numerical methods such as the finite element method (FEM) requires discretizing the problem spatially and temporally. The spatial discretization is usually represented using a mesh.
The quality of a mesh affects the accuracy of the numerical solution. 
However, it is often prohibitively expensive to solve the problem on a very high resolution mesh. 
Mesh adaptation is an advanced discretization method to tackle this problem. It increases the mesh resolution where the solution requires higher numerical accuracy, while decreasing the mesh resolution where unnecessary. Mesh adaptation methods can be generally divided into two categories: $h$-adaptation and $r$-adaptation. 
In $h$-adaptation, new mesh nodes are dynamically added to the regions where fine resolution is required. In $r$-adaptation (or \emph{mesh movement}), mesh nodes are only relocated or moved without changing the mesh topology. Compared to $h$-adaptation, $r$-adaptation has several attractive features. 
First, no extra mesh points are generated, which keeps the dimension of the linear system representing the discretized PDE unchanged.
In addition, fixed mesh connectivity can also make the structure of the stiffness matrix unchanged, which enables matrix pre-factorization that can accelerate the solution of the large linear systems encountered in FEM \citep{budd2009adaptivity}.
However, a key challenge for mesh movement methods is mesh tangling, whereby one or more elements become inverted. Mesh movement methods based on optimal transport theory
    ~\citep{Weller2016,McRae2018} can effectively prevent mesh tangling issues (see~\citep{clare2022multi} for a discussion on this), but require solving a Monge-Amp\`ere equation at each adaptation step, which is extremely time-inefficient.

AI for the solution of PDEs has been an emerging topic in recent years, and shows great potential in solving problems where traditional numerical PDE solvers struggle (e.g. high dimensional problems \citep{han2018solving,sheng2021pfnn}), or in accelerating the solution process by learning a neural operator from the parameterized description of a PDE problem to its corresponding solution \citep{li2020fourier,lu2019deeponet}. However, these methods still encounter fatal bottlenecks, such as the precision guarantee of the solution, data efficiency, generalization capability, and so on. These are fundamental limits of deep learning, but essential for the scientific computing scenarios.

A possible alternative is to perform learning based mesh adaptation, by which the traditional numerical PDE solvers can achieve better performance, while the time consumption of mesh adaptation is greatly reduced. There have been some works in this direction \citep{zhang2020meshingnet,yang2021reinforcement,huang2021machine,FIDKOWSKI2021109957,tingfan2021mesh,pfaff2020learning}. However, most previous methods target mesh generation \citep{zhang2020meshingnet} or mesh refinement \citep{yang2021reinforcement,huang2021machine,FIDKOWSKI2021109957}, instead of topology-invariant mesh movement as considered here. Moreover, the previous methods are not end-to-end, which means the neural networks are used to predict certain metrics, such as the local mesh density \citep{zhang2020meshingnet,huang2021machine} or the metric tensor \citep{FIDKOWSKI2021109957,tingfan2021mesh,pfaff2020learning}, which then have to be fed into a traditional mesher/remesher to obtain the mesh. Therefore the overall performance is bounded by the mesher/remesher, which is computational geometry based and hence not optimized for solving PDE problems. On the contrary, in our method, the adapted mesh is directly output by the neural network.

In this work, we propose to the best of our knowledge the first learning-based end-to-end mesh movement framework for PDE solvers. Taking the source term, the input field, and/or the PDE parameters as input features, the model deforms an initial mesh to the adapted mesh by mesh movement. In usage, the model can be applied to a class of PDEs without retraining. We design a Neural-Spline based model for mesh deformation. It is an invertible neural network and hence can avoid mesh tangling. Moreover, its mechanism naturally guarantees a hypercubic boundary can be maintained through the learnable mapping. We also design a graph attention network (GAT) based model for mesh deformation. The graph neural network can naturally describe domains with irregular shapes and embed the relevant information. We utilize the attention mechanism of the GAT model to guarantee each mesh node stays within its neighborhood.

Our main contributions can be listed as below:
\begin{enumerate}
\item We propose a learning-based end-to-end mesh movement framework for PDE solvers, which to the best of our knowledge is the first of its kind. Without interfering with the PDE solver, the models can achieve numerical error reduction similar to the traditional Monge-Amp\`ere method, while the mesh generation process is accelerated by two to three orders of magnitude.
\item A Neural-Spline based model and a GAT based model are proposed for mesh deformation. Besides generalization to different PDE parameters, source terms, input solution field, etc., the models are specifically designed to guarantee boundary consistency, alleviate mesh tangling, and generalize to different mesh densities.
\end{enumerate}

\section{Related Work}
\paragraph{Mesh movement method.}
Mesh movement methods include velocity-based and location-based methods.
In the work of \citep{Rai1983,Gnoffo1982,Farhat1998}, the mesh is moved according to attraction and repulsion pseudo-forces between nodes motivated by a spring model. The moving mesh finite element method \citep{Baines2005Moving} computes the solution and the mesh simultaneously by minimizing the residual of the PDEs written in a finite element form. 
As for location-based method, the moving mesh PDE (MMPDE) method \citep{huang2011} moves the mesh through the gradient flow equation of an adaptation functional.  In recent years, there has been a growing interest in optimally-transported $r$-adapted meshes \citep{budd2009,Weller2016,McRae2018,clare2022multi}. 

\paragraph{AI for PDE.}
To solve a PDE problem, neural networks can be used to represent the function to solve, and trained either with the residual loss of the PDE or using the variational principle \citep{raissi2019physics,yu2017deep}. Neural operators are proposed to learn an operator from the problem function to the solution function \citep{li2020fourier,lu2019deeponet}. There also exist mesh-based PDE solvers with deep learning. In \citep{pfaff2020learning}, a graph neural network with additional world edges are applied to predict dynamical systems, shown to be effective with a wide range of physical systems. In \citep{belbute2020combining}, a differentiable PDE solver is embedded in a neural network to help predict accurate solutions and also backpropagate the loss so that the input coarse mesh can be optimized.

\paragraph{AI for Meshing.}
AI methods have also been proposed for mesh generation, adaptation, and so on. MeshingNet \citep{zhang2020meshingnet} uses a neural network to learn the required local mesh density, which can then be provided to a Delaunay triangulation based mesh generator to generate high quality meshes. The optimal local mesh density is also learned in \citep{huang2021machine} for mesh refinement. On the other hand, the mesh refinement process is formulated as a reinforcement learning problem in \citep{yang2021reinforcement} to minimize the PDE solution error under given refinement budgets. The flow field is predicted by machine learning models to calculate the metric tensor so that the mesh can be optimized accordingly \citep{tingfan2021mesh}. The authors in \citep{FIDKOWSKI2021109957} focused on optimal anisotropic meshes by predicting the desired element aspect ratio. In \citep{pfaff2020learning}, the sizing field is predicted by a neural network for adaptive remeshing along with the system dynamics.



\section{Method}
    





\subsection{Problem Statement}

Meshing is the procedure to spatially discretize a PDE, which is necessary for most numerical methods to solve PDEs. Among these methods, the Finite Element Method (FEM) is one that has been widely used in various engineering fields. A high-quality mesh can significantly improve the accuracy-efficiency trade-off of PDE numerical methods, therefore mesh adaptation techniques have been well studied, which in major include two categories, mesh refinement and mesh movement. In this work, we apply our mesh movement method with FEM.

Consider a discretized PDE $\mathcal{P}$ defined on the domain $\Omega$ in $D$ dimensions. 
For mesh adaptation, we can define two domains: $\Omega_i$ for the initial mesh $\mathcal{T}^i$, and $\Omega_d$ for the adapted mesh $\mathcal{T}^d $. $\mathcal{T}^d$ should be adapted in response to the structure of the physical system under consideration. 
We denote mesh node positions in $\mathcal{T}^i$ by ${\boldsymbol{\xi}}$, and in $\mathcal{T}^d$ by ${\boldsymbol x}$. The adapted mesh $\mathcal{T}^d$ is the image of $\mathcal{T}^i$ under the mapping ${\boldsymbol x} = \mathcal{F}(\mathcal{X}, \boldsymbol{\xi} | \theta)$, where $\mathcal{X}$ is the input state of the PDE and $\theta$ represents the model parameters.
The solution on $\mathcal{T}^i$ is denoted as ${\boldsymbol u}^i = \mathcal{P}({\mathcal{X}}( {\boldsymbol \xi}))$ serving as a baseline and the solution ${\boldsymbol u}^d = \mathcal{P}({\mathcal{X}}( {\boldsymbol x}))$  on $\mathcal{T}^d$ is the optimized solution enhanced by mesh adaptation. The computational cost of traditional methods for mesh adaptation is often expensive. For some methods, the cost of mesh adaptation is comparable to or even more expensive than that of solving the underlying PDE problem, which is generally unacceptable. Therefore, our goal is to model the mapping $\mathcal{F}(\cdot|\theta)$ using learning based methods such that the mesh adaptation process can be greatly accelerated.

\begin{figure*}
    \centering
    \includegraphics[width=0.99\textwidth]{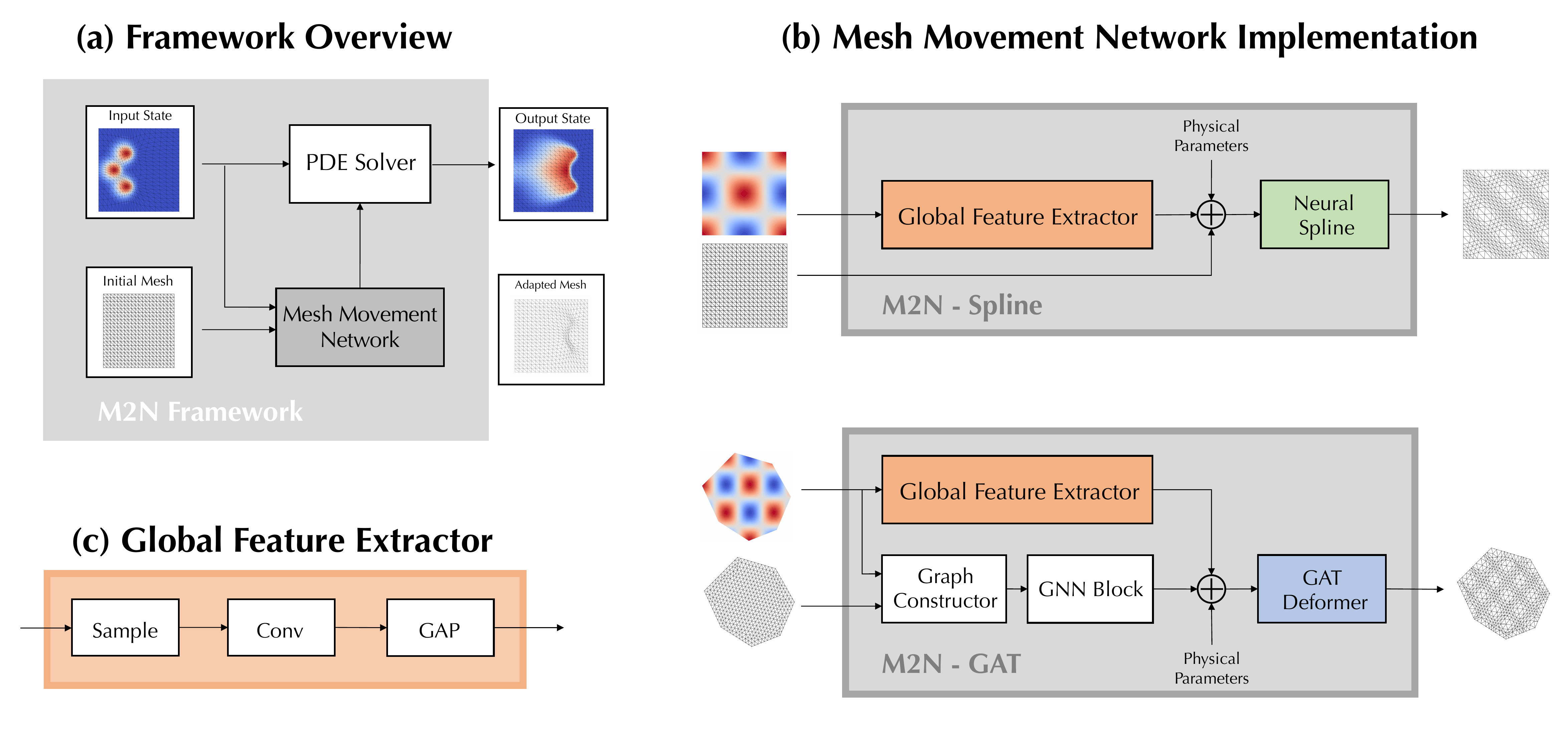}
    \caption{Proposed mesh movement networks framework. (a) Given an initial mesh and an input state, the mesh deformer outputs an adapted mesh, which is then fed to the PDE solver. (b) The implementations of the Neural-Spline and the GAT based mesh deformers are depicted. The GNN block indicates a Graph Neural Network block. (c) The Global Feature Extractor is composed of sampling on FEM function space, convolution and Global Average Pooling (GAP) layers.}
    \label{m2n_framework}
\end{figure*}

\subsection{Framework Overview}
The proposed framework is demonstrated in Figure \ref{m2n_framework}(a). For a class of PDEs, each of its specific sample $\mathcal{P}$ can be parameterized by PDE related information, such as source terms, boundary conditions, PDE parameters, solution fields, etc. We use $\mathcal{X}$ to denote such information and take it as the input of the neural network, so that the trained model can generalize to a class of PDEs.
The mesh movement network $\mathcal{F}(\cdot|\theta)$ deforms the initial mesh $\mathcal{T}^i$ and outputs an adapted mesh $\mathcal{T}^d$, given the input state $\mathcal{X}$. $\mathcal{T}^d$ is then fed into the PDE solver to help improve the accuracy of the solution. 
Under the M2N framework, we consider two implementations of the mesh movement network: a Neural-Spline based network and a GAT based network, which are denoted as M2N-Spline and M2N-GAT, respectively.

\subsection{Neural-Spline based Network}
Our Neural-Spline based network is mainly composed of two parts. One is a global feature extractor to extract features from the input state $\mathcal{X}$, the other is a neural spline deformer to output the adapted mesh $\mathcal{T}^d$.

\paragraph{Global Feature Extractor} 
We uniformly sample the input state $\mathcal{X}$ in the domain $\Omega_i$. If the domain is with an irregular boundary, sampling is performed inside its minimum bounding box. The values of the sampling points outside the domain boundary are set to zero. The sampled data are assembled as a state tensor $\mathcal{M}$. To keep the extracted feature invariant to the magnitude, the state tensor $\mathcal{M}$ is normalized by its maximum absolute value. After normalization, $\mathcal{M}$  is sent into the convolutional layers for feature extraction, whose output is further fed into a global average pooling layer to obtain a global embedding $\mathcal{E}$, which is invariant to the mesh resolution. The embedding $\mathcal{E}$ is then concatenated with physical parameters such as the mesh density and the PDE coefficient to get $\mathcal{I}$, which is fed into the deformer.

\paragraph{Neural-Spline based Deformer}
Normalizing flow models \citep{kobyzev2020normalizing} are proposed to learn invertible mappings. Neural spline \citep{durkan2019neural}, as a specific type of normalizing flows, transforms each dimension of the input with a differentiable monotone rational-quadratic spline function $\mathcal{S}(\cdot)$. More specifically speaking, for the $d$-th dimension of the node coordinates $\boldsymbol{\xi}$ of the initial mesh, the anchor points of the spline function are parameterized by the input features $\mathcal{I}$ and the coordinates of the other dimensions $\boldsymbol{\xi}_{-d}$, to get the output $\mathcal{S}_d(\boldsymbol{\xi}_{d}|\mathcal{I},\boldsymbol{\xi}_{-d})$. In a neural spline block, each dimension takes turns to be transformed, so that the node coordinates $\boldsymbol{\xi}$ of the initial mesh will be mapped to node coordinates $\boldsymbol x$ to assemble the adapted mesh $\mathcal{T}^d$.

Since the neural spline model is guaranteed to be an invertible mapping, mesh tangling can be alleviated. A specialty of the neural spline model is that, the end points of the spline function are fixed, therefore the intervals of the input and the output can be kept unchanged. In our case, this property is utilized to preserve the mesh with a hypercubic boundary (e.g. a rectangle in the 2-D case). Moreover, since the neural spline learns a continuous mapping, it can naturally be generalized to different mesh densities.


\subsection{GAT based Network}
Although the Neural-Spline based network is well-designed for meshes of hypercubic domains, it is difficult to extend to domains with more general boundaries. Therefore, we propose a more flexible model based on the Graph Attention Network (GAT) \citep{velivckovic2017graph}. It consists of a feature extractor and a GAT based mesh deformer.


\paragraph{Feature Extractor}
The feature extractor of the GAT-based network consists of two branches: one is the Global Feature Extractor which is similar to the Neural-Spline based network, the other is a Graph Neural Network (GNN) branch for local feature extracting.

For the GNN branch, a graph $G = (V, E)$ is constructed based on the initial mesh $\mathcal{T}^i$. Mesh nodes and mesh edges are transformed to graph nodes $V$ and bidirectional graph edges $E$. Then, we sample the values of the input state $\mathcal{X}$ at the position of each corresponding graph node $\boldsymbol v$ on the initial mesh. To capture the interaction between mesh nodes, the sampled values and mesh nodes geometry information are further processed by a GNN Block with message passing mechanism which can propagate the sampled physical quantities across the graph. To improve the generalization ability on mesh with different densities, the displacement vector ${\boldsymbol u}_{ij} = {\boldsymbol u}_i - {\boldsymbol u}_j$ and its norm $\| {\boldsymbol u}_{ij} \|$ are selected to describe the mesh density, where ${\boldsymbol u}_i$ denotes the position of mesh node $i$. They are encoded into mesh edges ${\boldsymbol e}_{ij} \in E$ as the relative edge features. These edge features are then updated by a Multi-Layer Perceptron (MLP) $f$ with three layers and the features of each mesh node ${\boldsymbol v}_i$ are updated by summing up the surrounding edge features: ${\boldsymbol e}_{ij}^\prime \leftarrow f({\boldsymbol e}_{ij}, {\boldsymbol v}_i, {\boldsymbol v}_j)$, ${\boldsymbol v}_i^\prime \leftarrow \sum_j {\boldsymbol e}_{ij}^\prime$. The features of each mesh node are concatenated with the extracted global feature ${\boldsymbol h}_g$ and physical parameters ${\boldsymbol p}$ (such as the viscosity coefficient in Burgers equation). The concatenated feature $H$ together with the mesh nodes position $\xi$ are sent into the GAT deformer for mesh movement. 

\begin{figure}
    \centering
    \includegraphics[width=0.7\textwidth]{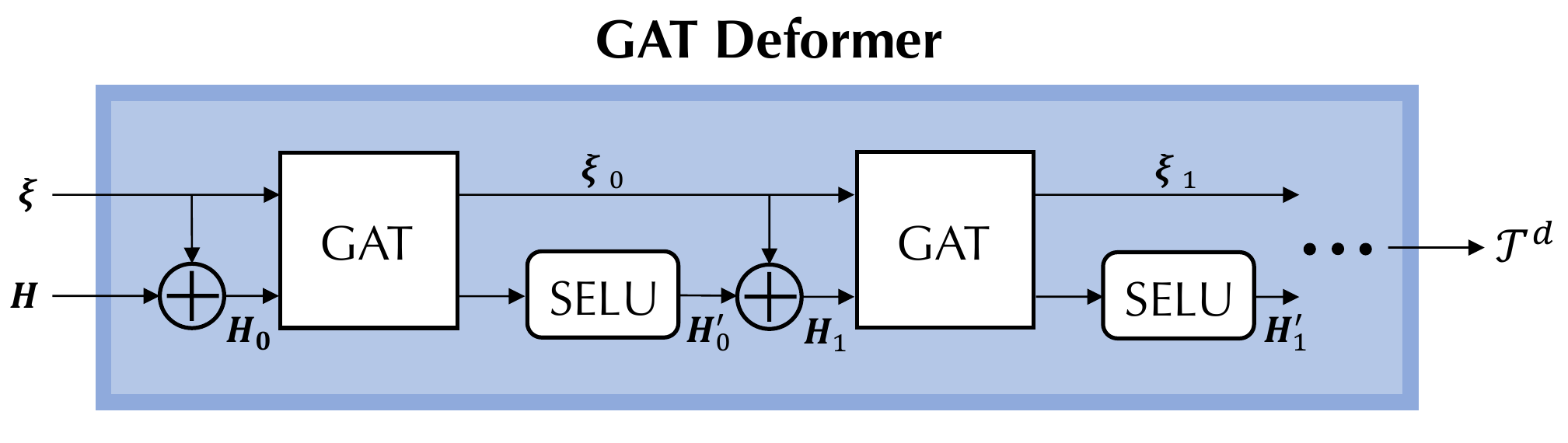}
    \caption{The implementation of the GAT Deformer. ${\boldsymbol \xi}_n$ and ${\boldsymbol  H}_n$ denote the mesh node positions and extracted features for the $n$th GAT block, while ${\boldsymbol  H}_n^\prime$ represent the corresponding outputs.}
    \label{fig:gat_deformer}
\end{figure}

\paragraph{GAT-based Deformer} As shown in Figure \ref{fig:gat_deformer}, the deformer consists of $N$ connected GAT blocks with SELU as the activation function. Each block takes two streams of inputs: mesh vertex positions ${\boldsymbol \xi}$ and extracted features ${\boldsymbol  H}$. To improve the learning efficiency, the intermediate output mesh nodes positions ${\boldsymbol \xi}_n$ are appended to the corresponding output features ${\boldsymbol  H}_n^\prime$ as the part of input features ${\boldsymbol  H}_{n+1}$ for next block. The GAT based deformer can naturally handle the problem of mesh tangling: during the deformation process, the movement of each mesh node is confined inside the convex hull composed of its 1-ring neighbors, which can effectively alleviate mesh tangling. 
To keep the boundary consistent before and after the deformation, the nodes on the boundary are restricted to move along the domain edge.

\section{Experiment}



We evaluate our proposed models against two baseline models using error reduction ratio, mesh generation speed, tangling avoidance, as well as generalization capability to different PDE problems and mesh densities. The experiments are conducted on the stationary and linear Poisson's equation in both a square domain and an irregular heptagonal domain, and the time-dependent non-linear Burgers' equation, where the supervised optimized meshes are generated with the traditional Monge-Ampère (MA) method.

\subsection{Experiment Setup}
\begin{figure}
\centering
    \begin{subfigure}{0.3\textwidth}
        \includegraphics[width=\textwidth]{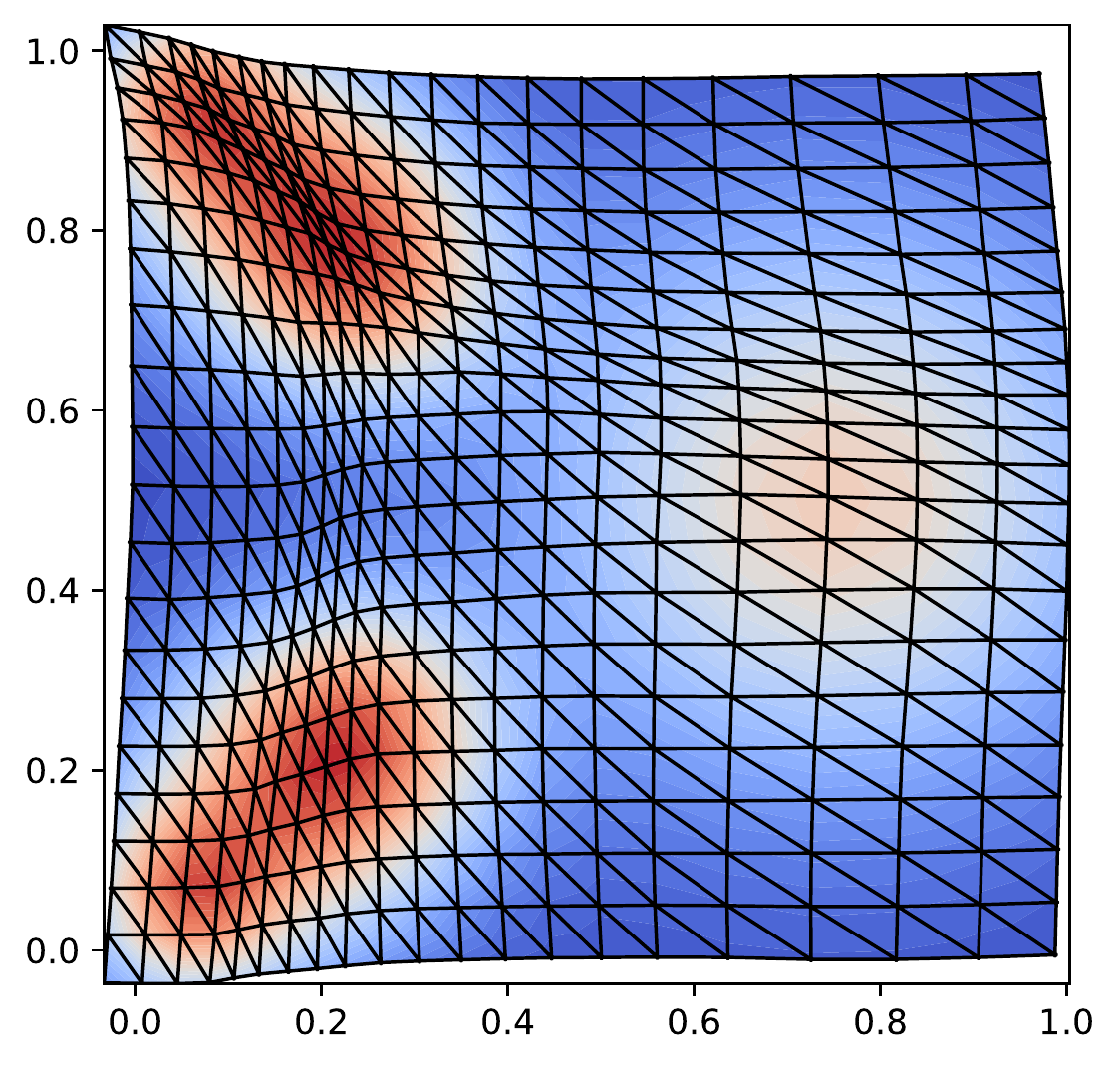}
        \caption{Before clipping}
    \end{subfigure}
    \quad
    \begin{subfigure}{0.3\textwidth}
        \includegraphics[width=\textwidth]{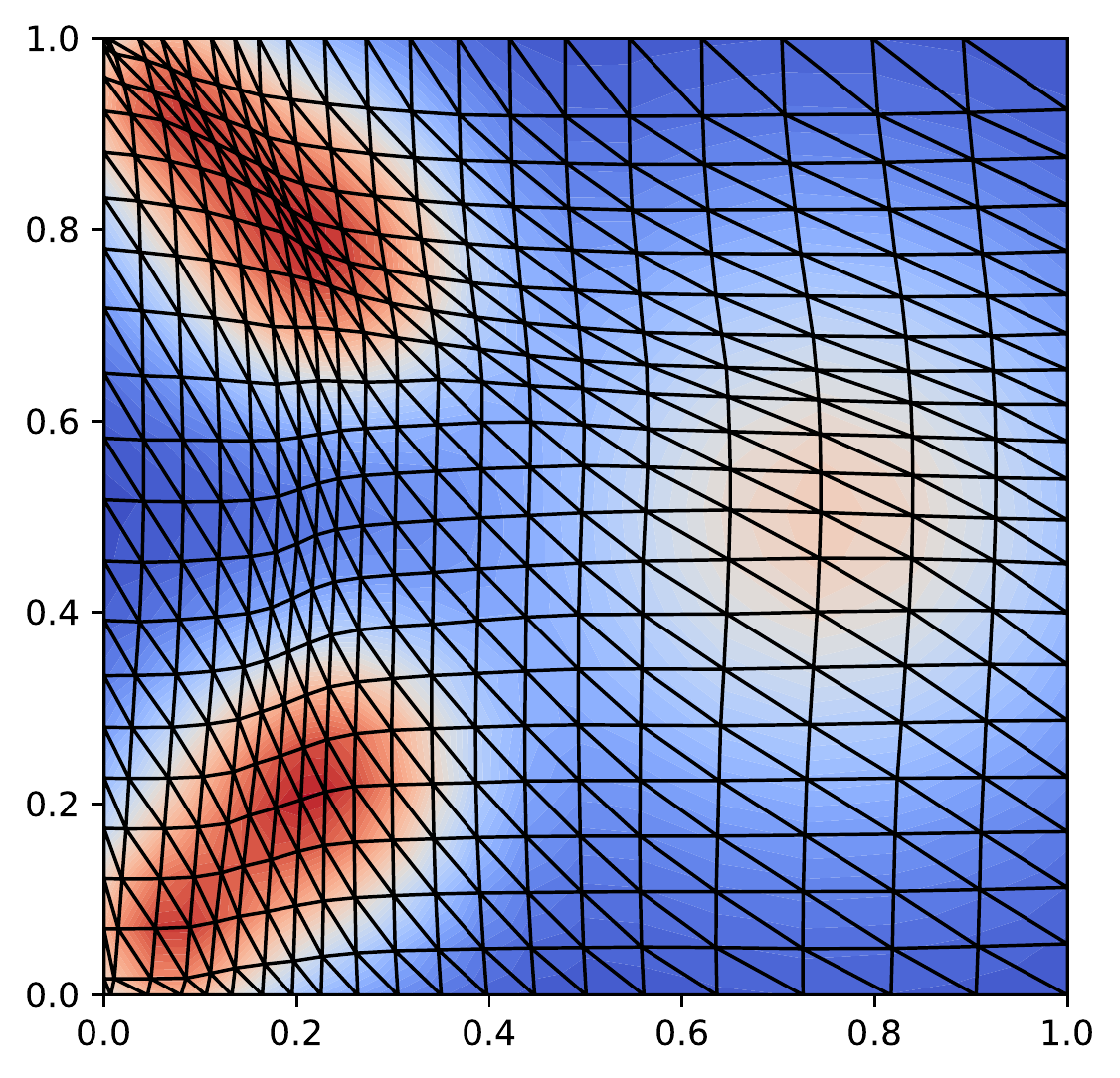}
        \caption{After clipping}
    \end{subfigure}
    \hfill
    \caption{Illustration of the mesh movement results of the baseline models before and after clipping.}
    \label{fig:clipped_compare}
\end{figure}

\paragraph{Comparison Baseline.} As mentioned in Section \ref{section:intro}, there is no similar previous work to compare results against. Therefore, we compare our proposed models with two baselines that can be interpreted as ablated versions of our proposed models. The model MLP-Deform-Clip replaces the neural-spline block with MLP. The model GAT-Deform-Clip replaces the GAT-deformer with the ordinary GAT block. Instead of learning the positions of mesh nodes, we set the learning target as the mesh nodes displacement for the baseline models because it gives better performance. In order to enforce that baseline models can also preserve boundary consistency, the nodes moved out of the boundary will be pulled back into the domain, and the displacement component perpendicular to the boundary is clipped for the nodes which are supposed to stay on the boundary, as shown in Figure \ref{fig:clipped_compare}. However, there is no trivial way to alleviate mesh tangling.

\paragraph{Monitor Function.} In all experiments, we use the Monge-Amp\`ere (MA) method as described in \citep{McRae2018} to generate the target deformed mesh. For the MA method, the mesh is equidistributed with respect to a user-defined monitor function.
The monitor function provides a concept of `mesh density' and controls the sense in which errors are equidistributed by the mesh adaptation algorithm.
Therefore, the choice of monitor function greatly impacts the geometry of the deformed mesh and should be chosen with care. In this paper, we define the monitor function as
\begin{equation}
    m = 1 + \alpha \frac{(u-u_{exact})^2}{\max_{x,y}(u-u_{exact})^2} + \beta \frac{\| H(u) \|_{F}}{\max_{x,y}\| H(u) \|_{F}},
\end{equation}
where $u$ is the numerical solution on the current mesh, $u_{exact}$ is the numerical solution on a uniformly refined mesh or analytical solution, $\| H(u) \|_{F}$ represents the Frobenius norm of the Hessian matrix of the solution $u$.


\paragraph{Implementation Details.}
For all experiments, meshes are evaluated using PDE solutions obtained from the open-source FEM solver Firedrake \citep{Rathgeber2016}. The neural network models are implemented with PyTorch \citep{paszke2019pytorch}, and the graph neural networks are implemented with PyTorch Geometric (PyG) \citep{Fey/Lenssen/2019}. We use the $\ell^1$ loss to train the models, i.e., $L(\mathbf{\xi},\bar{\mathbf{x}}; \theta)=\| F_{\theta}(\mathbf{\xi})- \bar{\mathbf{x}} \|_1$, where $\bar{\mathbf{x}}$ denotes the positions of target mesh nodes. The models are trained with the Adam optimizer \citep{kingma2014adam}, using a mini-batch size of 6 meshes. Each model is trained three times with different random seeds. 

\paragraph{Evaluation Metric.} We evaluate the performance of different models by calculating the average error reduction ratio of the deformed mesh generated by different models.
The quality of the mesh is evaluated by calculating the percentage reduction of the discretization error. The ground truth is provided either by the numerical solutions on meshes with very high resolution or the analytical solutions by proper dataset generation. We calculate the percentage of error reduction with the formula $(e_{\text{initial}} - e_{\text{adapted}}) / e_{\text{initial}}$, where $e_{\text{initial}}$ is the error norm of the solution solved on initial uniform mesh compared with the ground truth. Similarly, $e_{\text{adapted}}$ is the error norm of the solution obtained from adapted deformed mesh.

\subsection{Poisson's Equation}
Poisson's equation is a second order, linear, stationary PDE, which is widely used in electrostatics and thermodynamics, amongst other fields. We consider the 2-D case with a Dirichlet boundary condition:
\begin{align}
\begin{split}
    -\nabla \cdot \nabla u(x, y) &= f(x, y), \quad\:\: (x,y) \in \Omega, \\
    u(x, y) &= u_0(x, y),  \quad  (x,y) \in \partial\Omega.
\end{split}
\end{align}
For both the square and heptagonal domain experiments, we generate analytical $u$ samples from a mixed Gaussian distribution, which are fed into the Poisson's equation to obtain the corresponding source terms $f$ and boundary conditions $u_0$ as the problem samples, whereas the $u$ functions serve as the ground truth.


\paragraph{Square Domain}
In this experiment, we train the models on cases with mesh resolution of 15x15 and 20x20, each with 275 samples. To evaluate the models and how well they generalize to different mesh resolutions, we test on cases with mesh resolution from 12x12 to 23x23, each with 125 samples. Moreover, we deliberately set the optimal mesh movement to be drastic, in order to test how well different models can handle mesh tangling.

\begin{table}[tb]
\caption{Performance summary of the Poisson's equation problem on the square domain.}
\centering
 \begin{tabular}{lrrr}
  \toprule
  Method & Error Reduction (\%) & Time (ms) & Element Inversion (\%) \\
  \midrule
  MA (traditional) & 23.11  & 5220.99 & 0.00 \\
  M2N-Spline & \textbf{20.82 $\pm$ 0.35} & 5.55 $\pm$ 0.01 & \textbf{0.00} \\
  MLP-Deform-Clip & 16.74 $\pm$ 0.90 & \textbf{3.02 $\pm$ 0.03} & 1.60\\
  M2N-GAT & \textbf{20.38 $\pm$ 0.51} & 9.09 $\pm$ 0.02 & \textbf{0.00} \\
  GAT-Deform-Clip & 19.95 $\pm$ 0.38 & 10.41 $\pm$ 0.05 & 3.11 \\
  \bottomrule
 \end{tabular}
 \label{tab:table_squared_poisson}
\end{table}


The quantitative results are summarized in Table \ref{tab:table_squared_poisson}. The proposed models, M2N-Spline and M2N-GAT, can achieve similar error reduction  compared to the traditional MA method, while the mesh generation speed is two to three orders of magnitude faster. In addition, although the proposed models perform only marginally better than the two baseline models in error reduction and time, they are proven very effective to keep the mesh untangled. In comparison, the baseline models suffer from mesh tangling.

\begin{figure}
\centering
\begin{minipage}{0.25\linewidth}
    \begin{subfigure}{1\textwidth}
    \includegraphics[width=\textwidth]{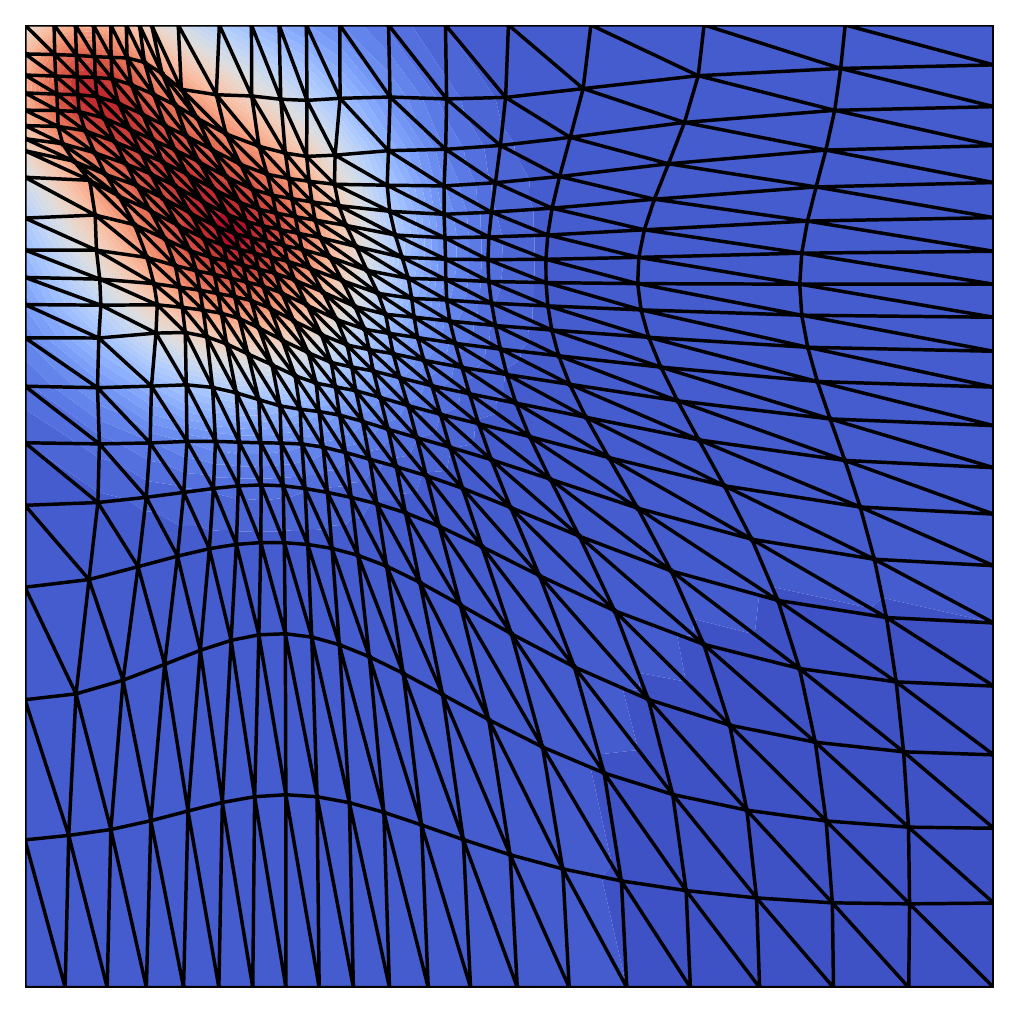}
    \caption{MA (37.55\%)}
    \end{subfigure}
\end{minipage}
\quad
\begin{minipage}{0.5\linewidth}
    \begin{subfigure}{0.45\textwidth}
        \includegraphics[width=\textwidth]{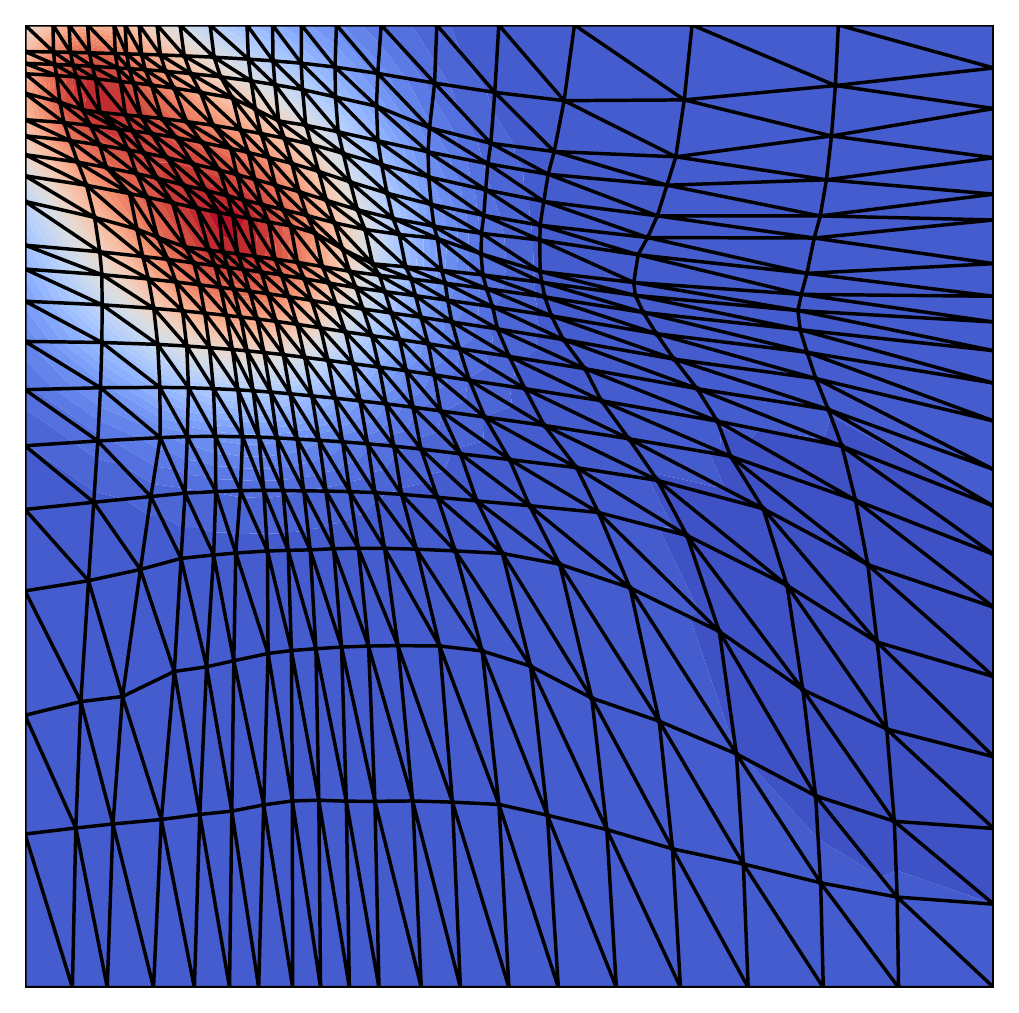}
        \caption{M2N-Spline (34.3\%)}
    \end{subfigure}
\hfill
    \begin{subfigure}{0.45\textwidth}
        \includegraphics[width=\textwidth]{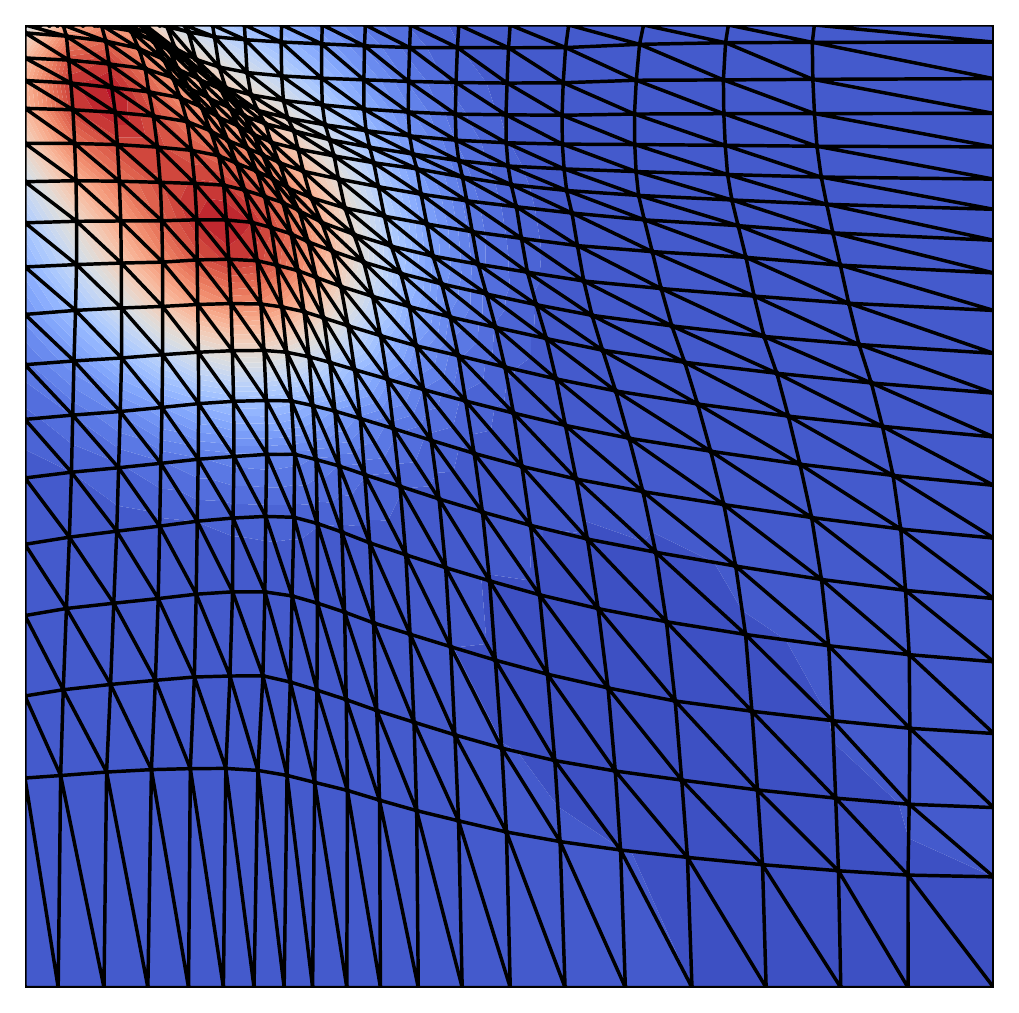}
        \caption{MLP-Deform-Clip (illegal)}
    \end{subfigure}
\\
    \begin{subfigure}{0.45\textwidth}
        \includegraphics[width=\textwidth]{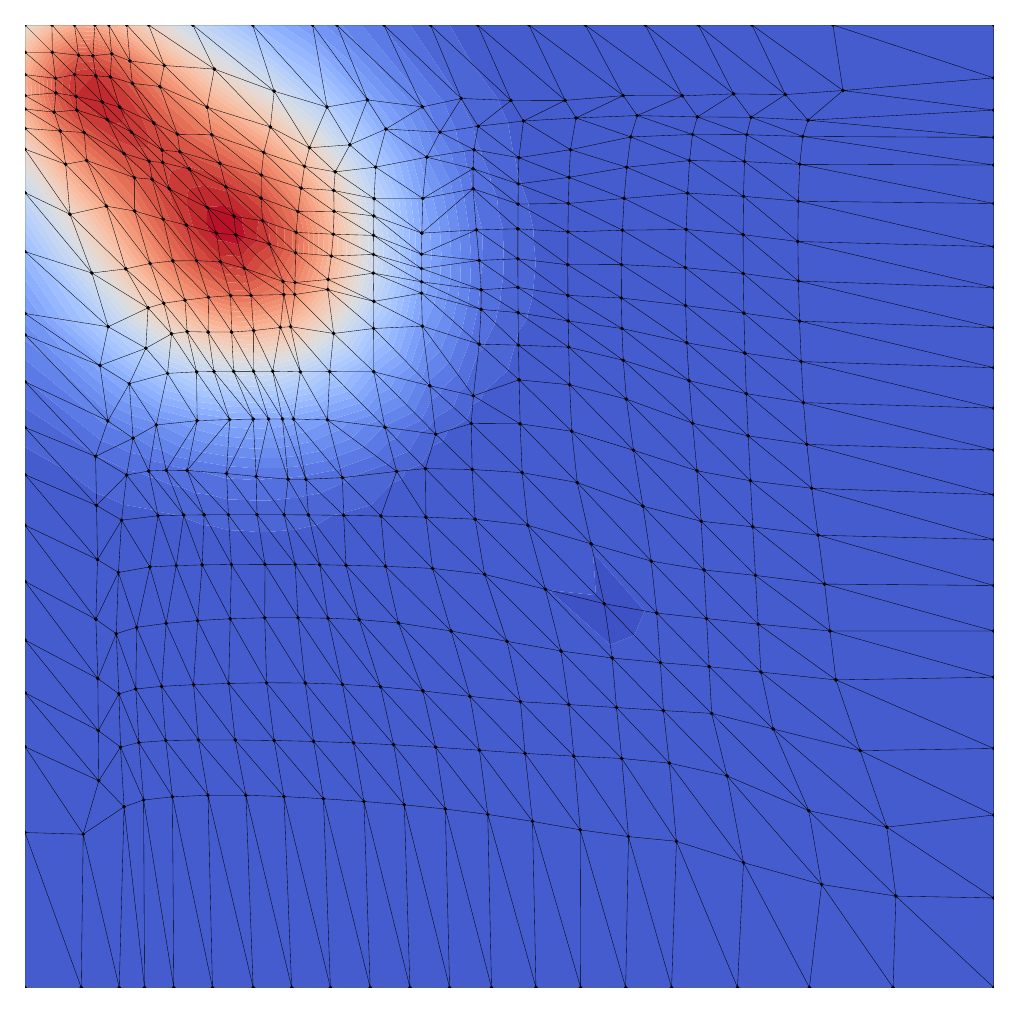}
        \caption{M2N-GAT (32.05\%)}
    \end{subfigure}
\hfill
    \begin{subfigure}{0.45\textwidth}
        \includegraphics[width=\textwidth]{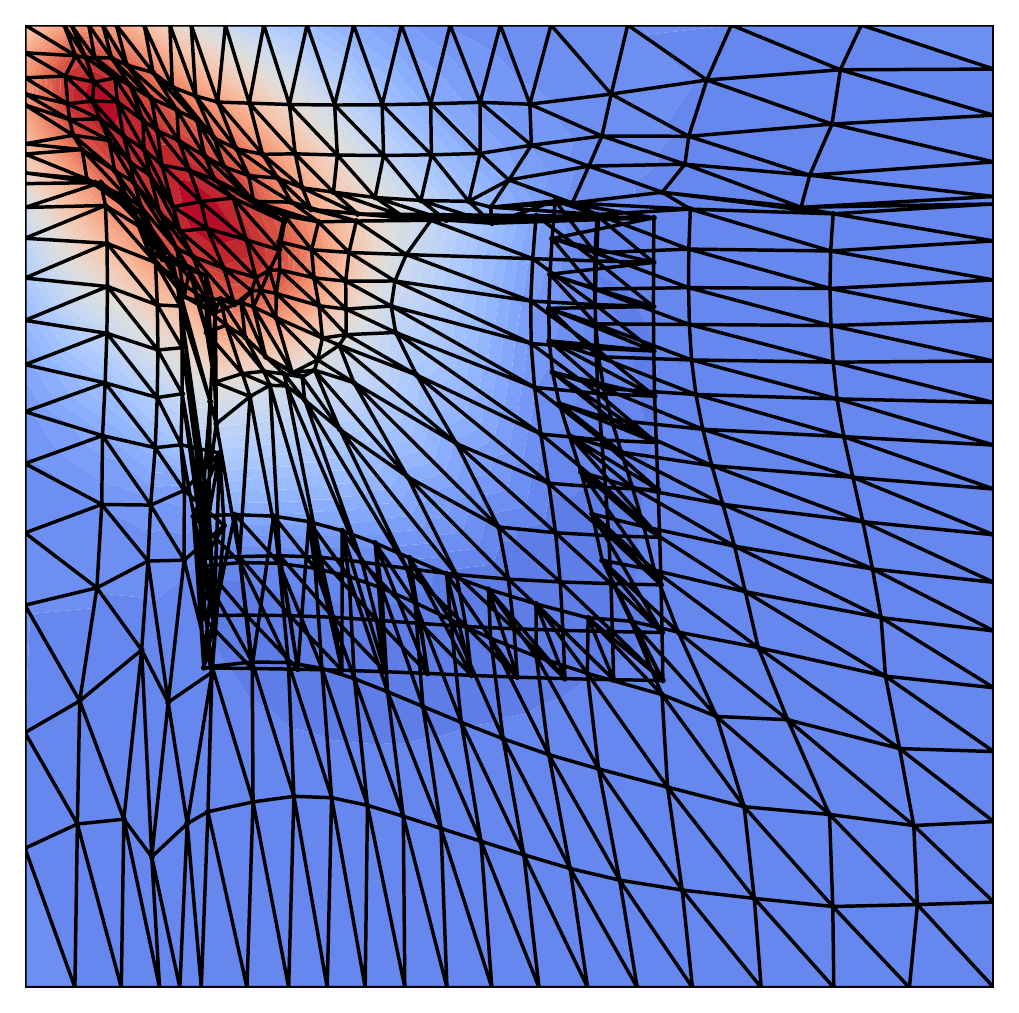}
        \caption{GAT-Deform-Clip (illegal)}
    \end{subfigure}
\end{minipage}
\caption{Comparison of mesh movement for an example problem of the Poisson's equation on the square domain. The percentage values in the parentheses are the error reduction ratios.}
\label{fig:large_deformation}
\end{figure}

An example is given in Figure \ref{fig:large_deformation}, where the mesh density in the upper left corner needs to be very high (shown in Figure \ref{fig:large_deformation}(a)). It is demonstrated that, M2N-Spline is more flexible for the cases where the overall mesh deformation is required to be large. On the other hand, for M2N-GAT, because of the constrained movement in each layer to alleviate mesh tangling and the finite layer numbers, it does not learn as well in such scenarios.

\paragraph{Irregular Heptagonal Domain} To evaluate the performance of different models for more general domain shapes, we conduct an experiment using Poisson's equation in an irregular heptagonal domain. The models are trained at mesh densities of 13, 16, 19, and 22, each with 320 samples, and tested on mesh densities from 12 to 23, each with 80 samples, to evaluate the performance and generalization capability of the models.

\begin{table}[tb]
\caption{Performance summary of the Poisson's equation problem on the heptagonal domain.}
\centering
 \begin{tabular}{lrrr}
  \toprule
  Method & Error Reduction (\%) & Time (ms) & Element Inversion (\%) \\
  \midrule
   MA (traditional) & 26.57  & 7329.78 & 0.00 \\
   M2N-Spline & 16.15 $\pm$ 0.40 & 5.97 $\pm$ 0.05 & 0.28 \\
   MLP-Deform-Clip & 16.49 $\pm$ 0.46 & \textbf{2.60 $\pm$ 0.02} & 0.56 \\
   M2N-GAT & \textbf{22.39 $\pm$ 0.27} & 9.41 $\pm$ 0.03 & \textbf{0.00} \\
   GAT-Deform-Clip & \textbf{22.40 $\pm$ 0.47} & 10.81 $\pm$ 0.07 & 0.68 \\
  \bottomrule
 \end{tabular}
\label{tab:table_polygon_poisson}
\end{table}

\begin{figure}
\centering
\begin{minipage}{0.25\linewidth}
    \begin{subfigure}{1\textwidth}
    \includegraphics[width=\textwidth]{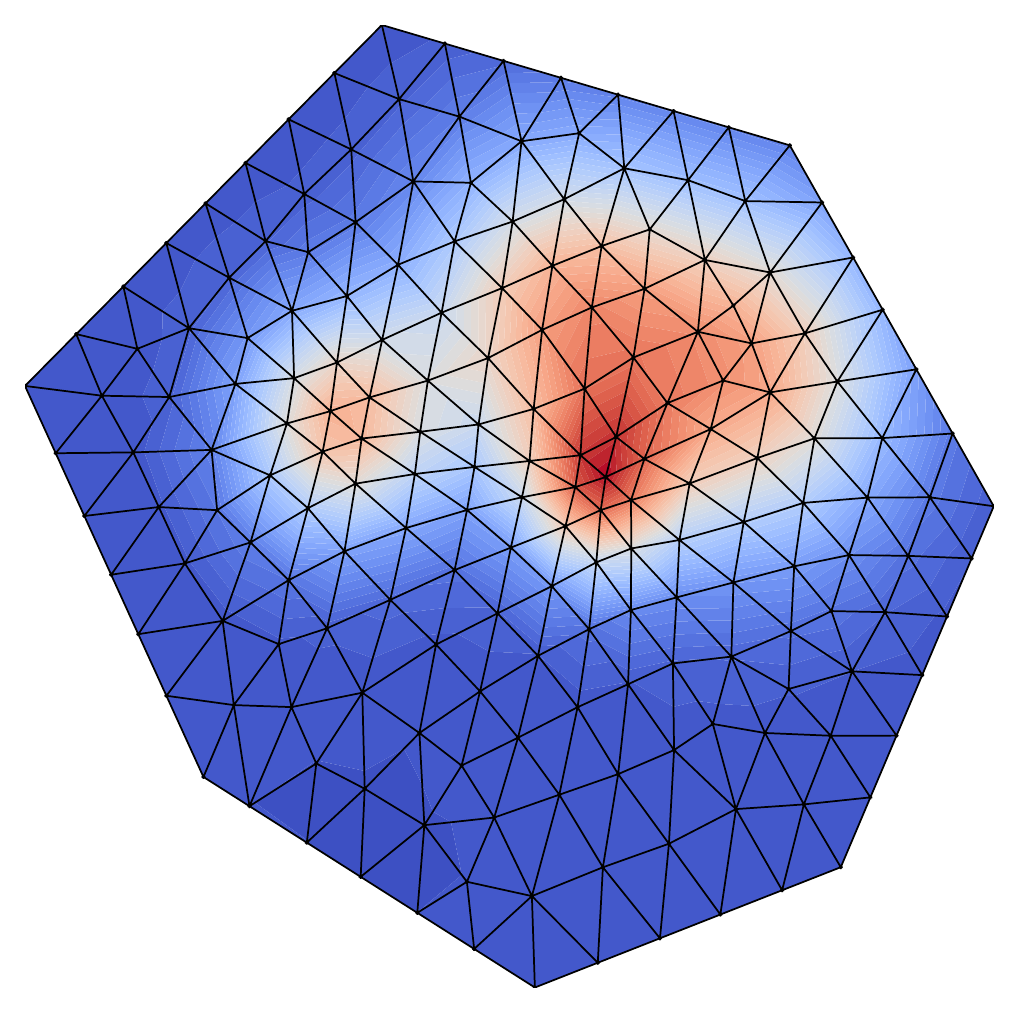}
    \caption{MA (33.35\%)}
    \end{subfigure}
\end{minipage}
\quad
\begin{minipage}{0.5\linewidth}
    \begin{subfigure}{0.45\textwidth}
        \includegraphics[width=\textwidth]{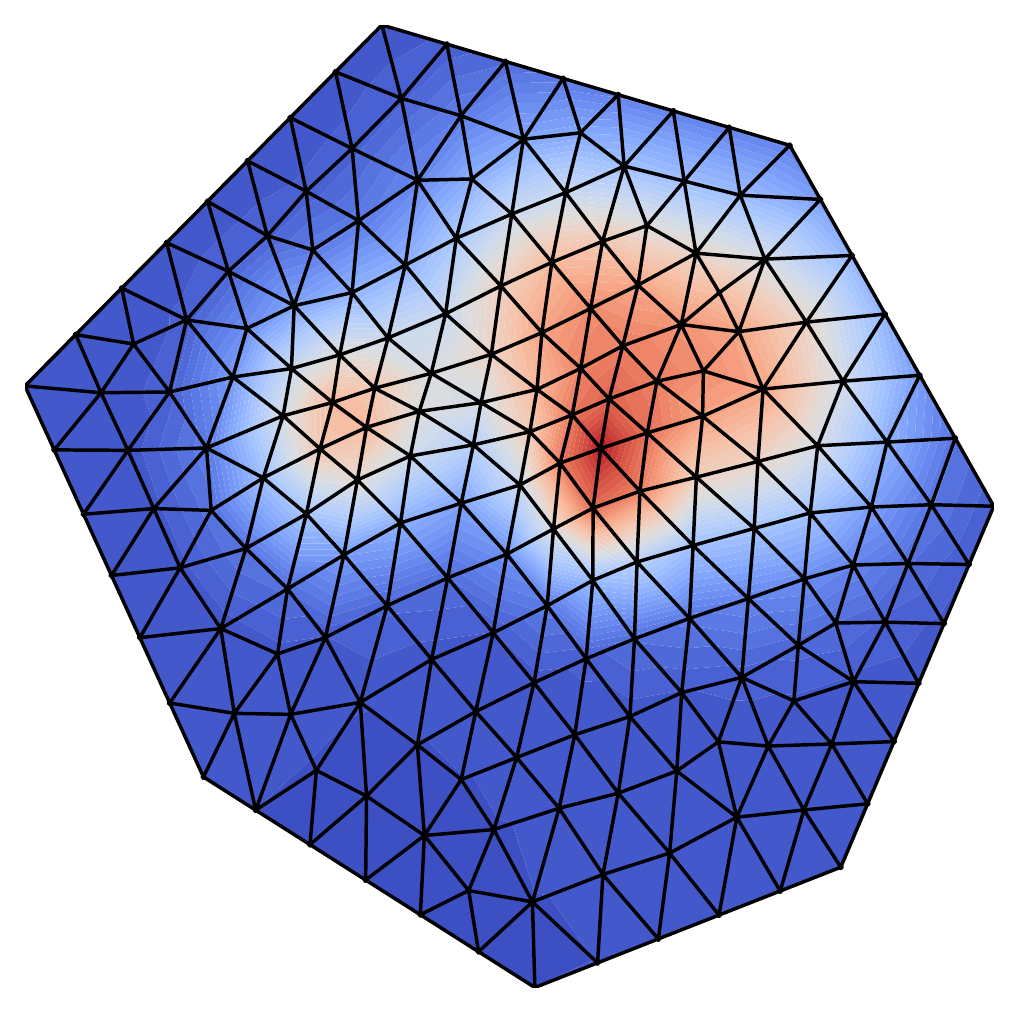}
        \caption{M2N-Spline (14.98\%)}
    \end{subfigure}
\hfill
    \begin{subfigure}{0.45\textwidth}
        \includegraphics[width=\textwidth]{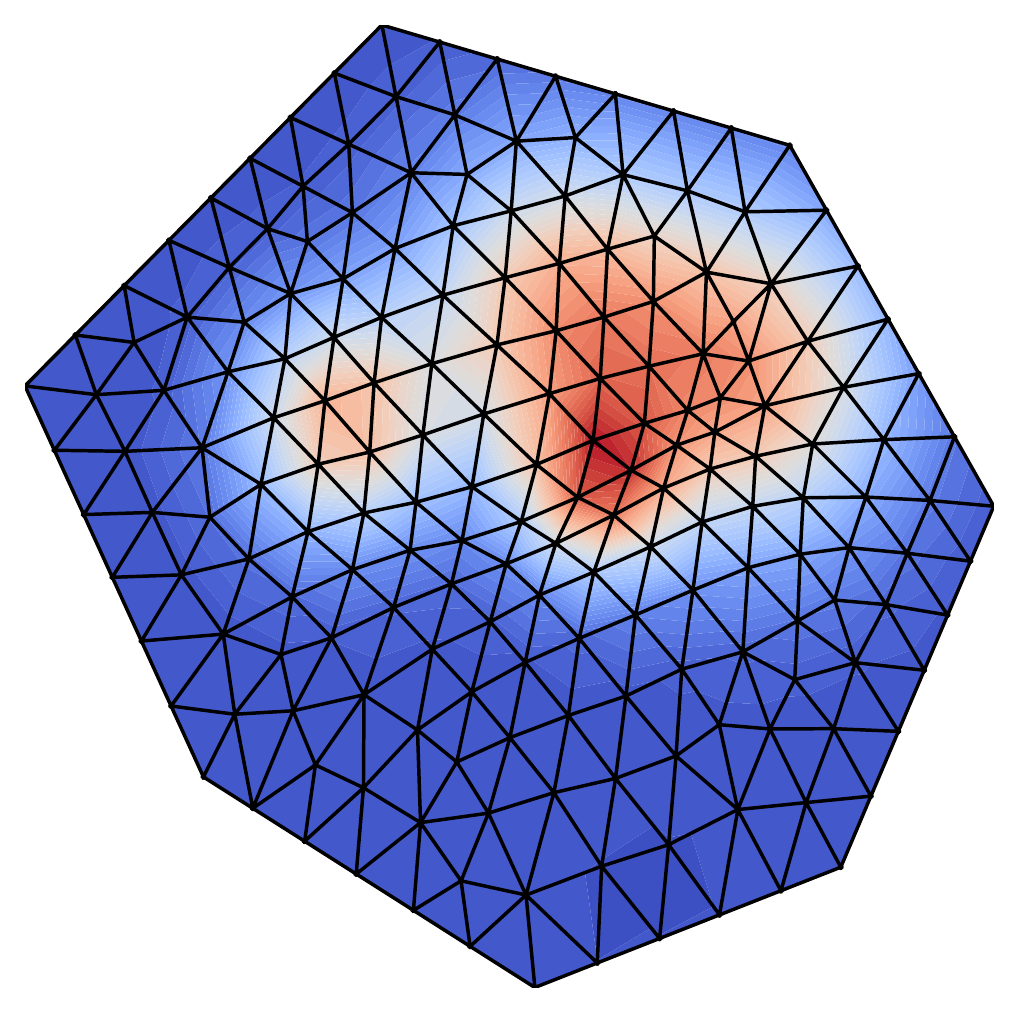}
        \caption{MLP-Deform-Clip (10.40\%)}
    \end{subfigure}
\\
    \begin{subfigure}{0.45\textwidth}
        \includegraphics[width=\textwidth]{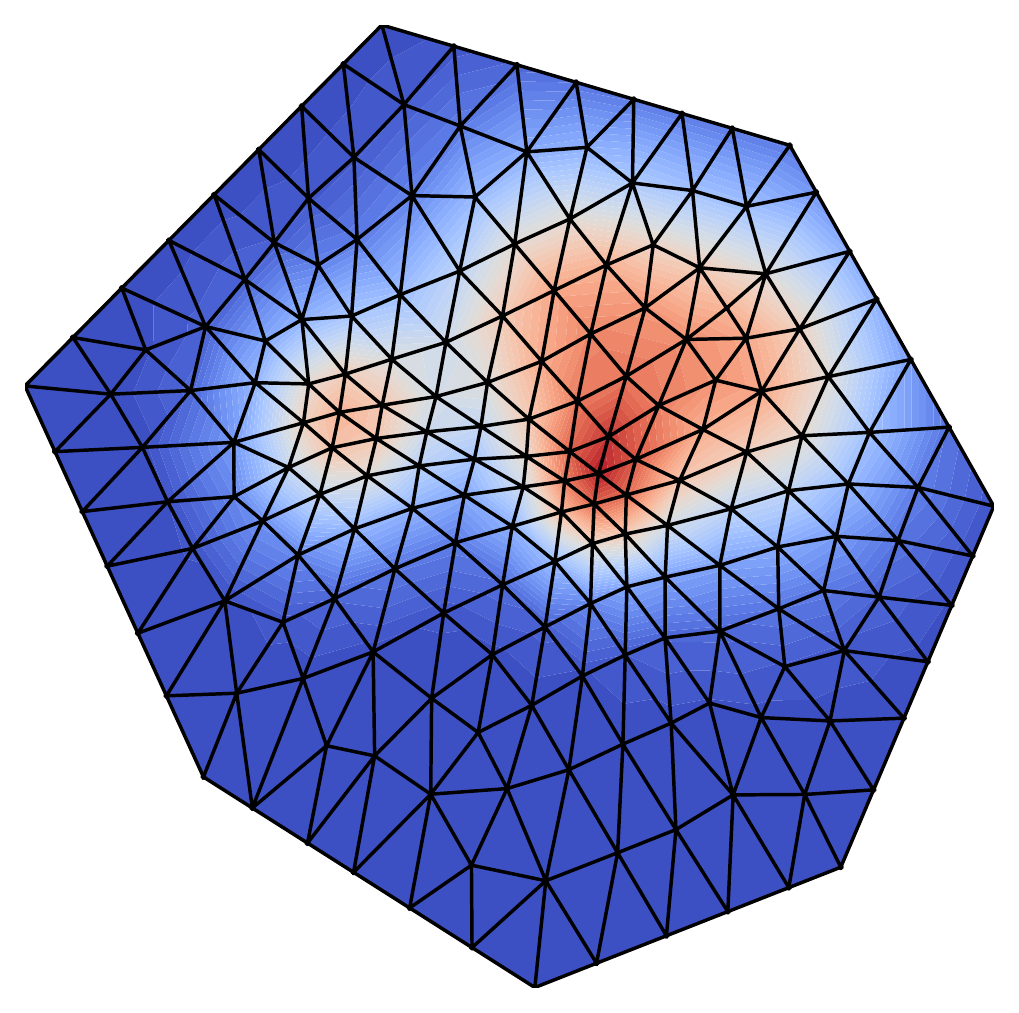}
        \caption{M2N-GAT (28.33\%)}
    \end{subfigure}
\hfill
    \begin{subfigure}{0.45\textwidth}
        \includegraphics[width=\textwidth]{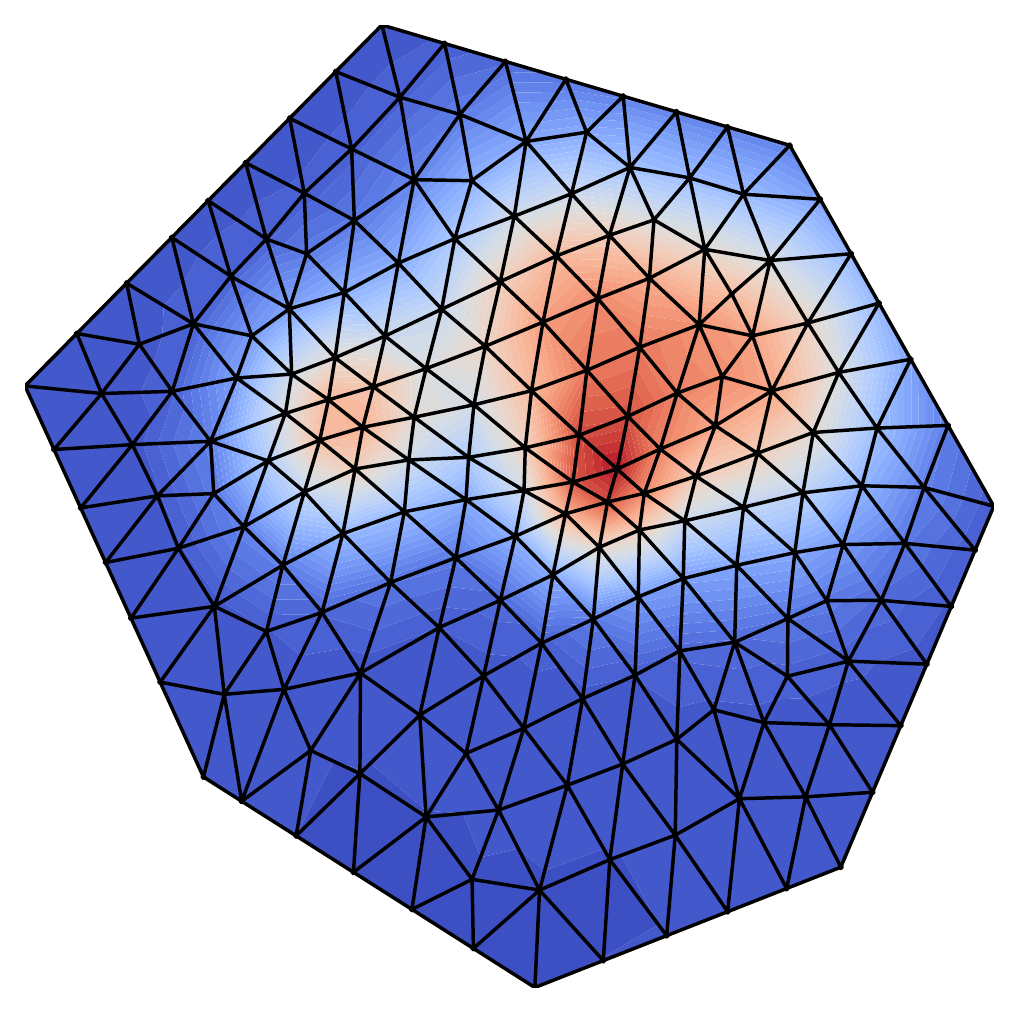}
        \caption{GAT-Deform-Clip (27.16\%)}
    \end{subfigure}
\end{minipage}
\caption{Comparison of mesh movement for an example problem of the Poisson's equation on the heptagonal domain. The percentage values in the parentheses are the error reduction ratios.}
\label{fig:polygon_mesh_demo}
\end{figure}

The results are summarized in Table \ref{tab:table_polygon_poisson}. It can be seen that all deep learning models can perform mesh movement around two to three orders of magnitude faster than the traditional MA method. Two GNN-based models perform better than the other two models, with the error reduction ratio comparable to the MA method. This is because the GNN-based models can naturally embed the information of the entire irregular domain into the network, and there are extra local feature extractors in the model. On the contrary, both the M2N-Spline and the MLP-Deform-Clip models are point-to-point mappings with only the global feature extractor, hence lack such capabilities. The results of all methods for an example problem are shown in Figure \ref{fig:polygon_mesh_demo}, from which we can see that the GNN-based models can better capture the delicate local structure where mesh resolution needs to be increased. Although mesh tangling still will not occur for M2N-GAT, it happens for M2N-Spline, because for the M2N-Spline model, its guarantee only works for hypercubic boundaries (rectangle in the 2-D case).

\subsection{Burgers' Equation}
The viscous Burgers' equation is a non-linear, time-dependent PDE describing advection and diffusion processes in fluids. Here we consider the 2-D Burgers' equation,
\begin{align}
\begin{split}
    \frac{\partial u}{\partial t}+(u \cdot \nabla) u-\nu \nabla^{2} u &= 0, \quad (x,y) \in \Omega,\\
    (n \cdot \nabla) u &= 0, \quad (x,y) \in \partial\Omega,
\end{split}
\end{align}
where constant scalar $\nu$ represents viscosity and $u$ is the velocity vector field obeying this PDE. We define the problem on the unit square domain $\Omega=[0, 1]^2$.

\begin{table}[tb]
\caption{Performance summary of the Burgers' equation problem.}
\centering
 \begin{tabular}{lrrr}
  \toprule
  Method & Error Reduction (\%) & Time (ms) & Element Inversion (\%) \\
  \midrule
   MA (traditional) & 60.24  & 81590.64 & 0.00 \\
   M2N-Spline & 48.92 $\pm$ 1.33 & 5.54 $\pm$ 0.02 & \textbf{0.00} \\
   MLP-Deform-Clip & 43.53 $\pm$ 1.91 & \textbf{2.92 $\pm$ 0.01} & \textbf{0.00} \\
   M2N-GAT & \textbf{57.75 $\pm$ 0.68} & 8.93 $\pm$ 0.01 & \textbf{0.00} \\
   GAT-Deform-Clip & 51.69 $\pm$ 4.01 & 10.41 $\pm$ 0.01 & 0.53 \\
  \bottomrule
 \end{tabular}
\label{tab:table_squared_burgers}
\end{table}

In this experiment, we still train the models on cases with mesh
resolution of 15x15 and 20x20, each with 9 trajectories and 60 timesteps per trajectory, with different viscosity coefficients. To evaluate the models and how well they can generalize to different mesh resolutions, we test on cases with mesh resolution from 11x11 to 24x24, each with 8 trajectories and 60 timesteps per trajectory. 

\begin{figure}
\centering
    \begin{subfigure}{0.2\textwidth}
        \includegraphics[width=\textwidth]{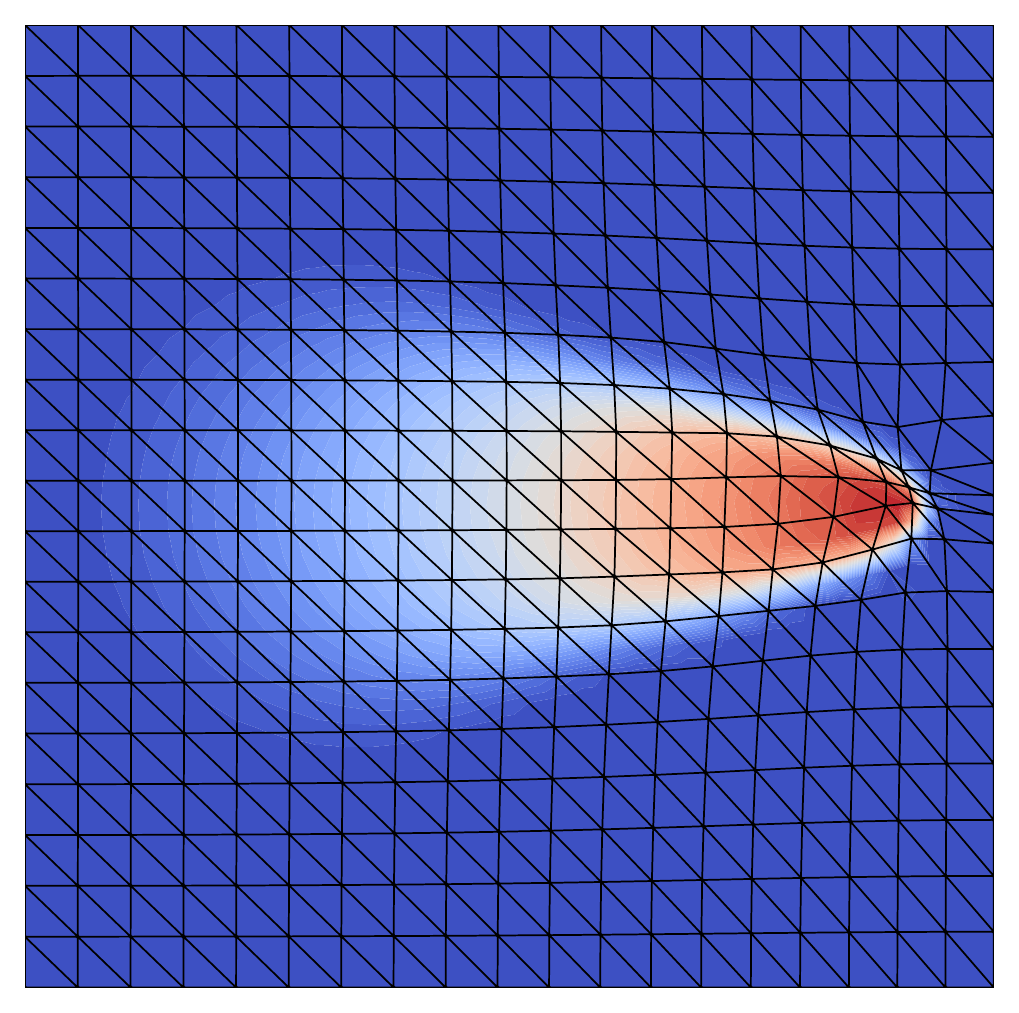}
    \end{subfigure}
\quad
    \begin{subfigure}{0.2\textwidth}
        \includegraphics[width=\textwidth]{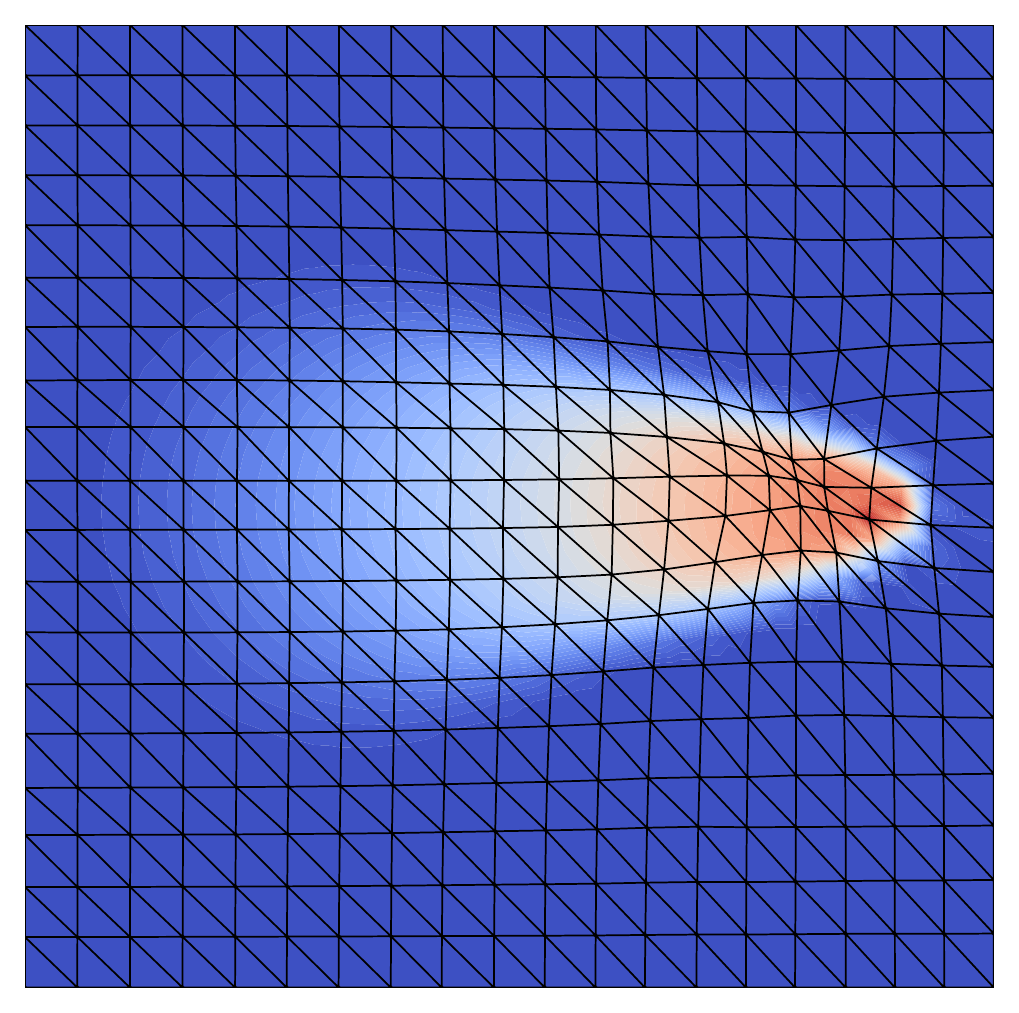}
    \end{subfigure}
\quad
    \begin{subfigure}{0.2\textwidth}
         \includegraphics[width=\textwidth]{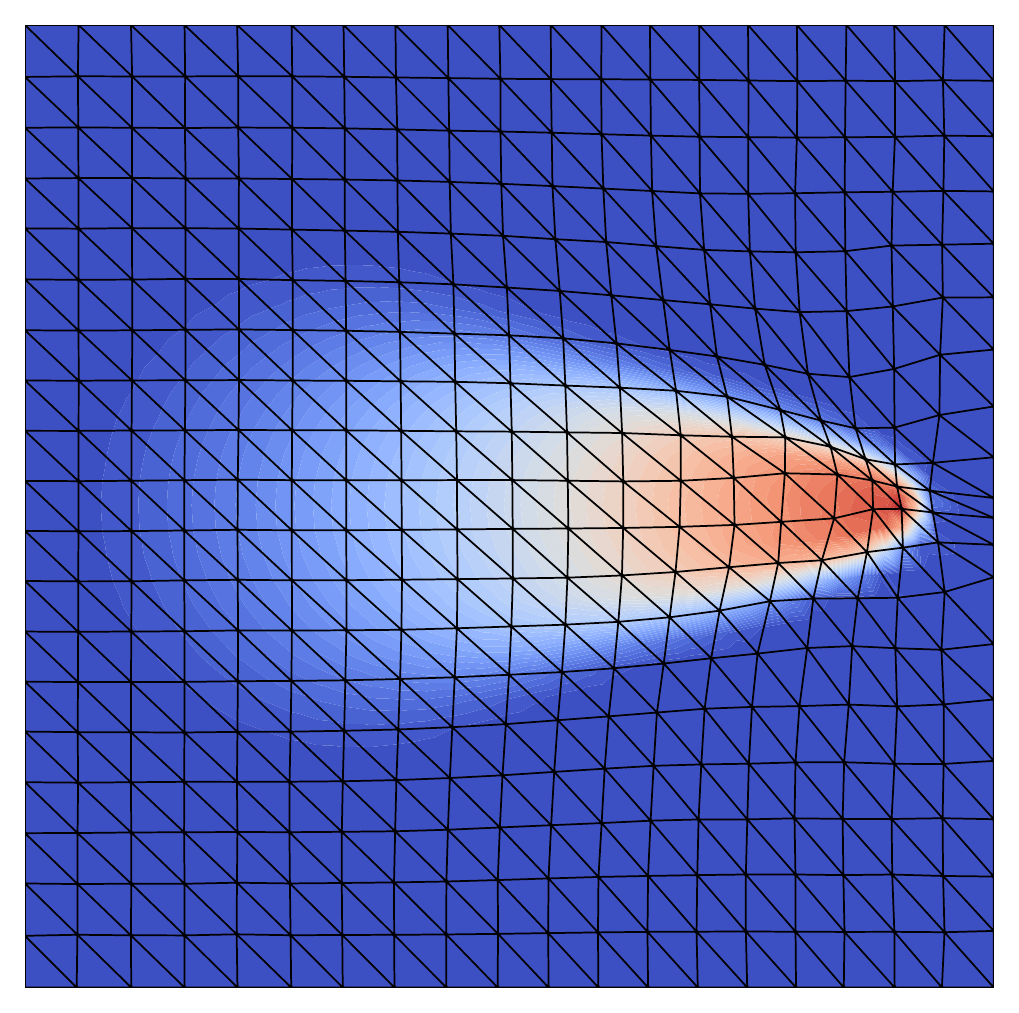}
    \end{subfigure}
\\
      \begin{subfigure}{0.2\textwidth}
        \includegraphics[width=\textwidth]{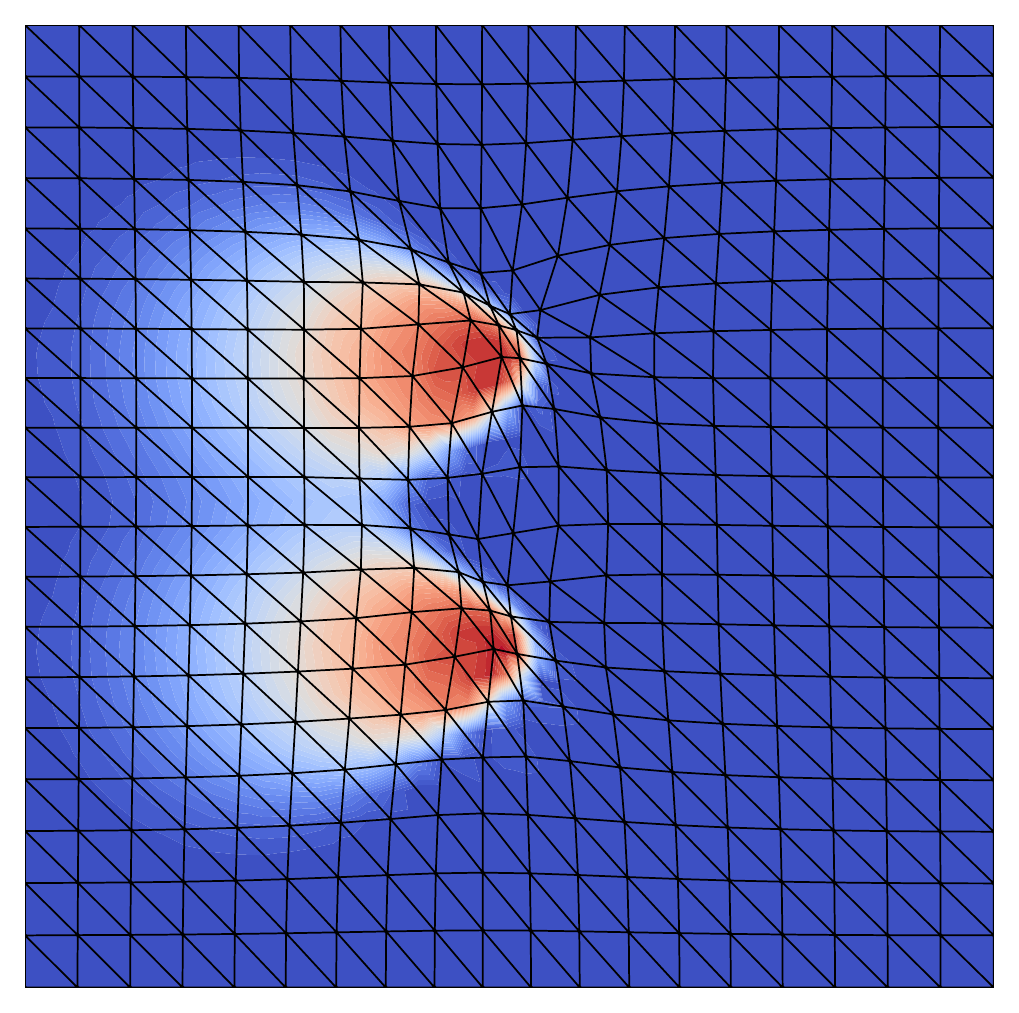}
    \end{subfigure}
\quad
    \begin{subfigure}{0.2\textwidth}
        \includegraphics[width=\textwidth]{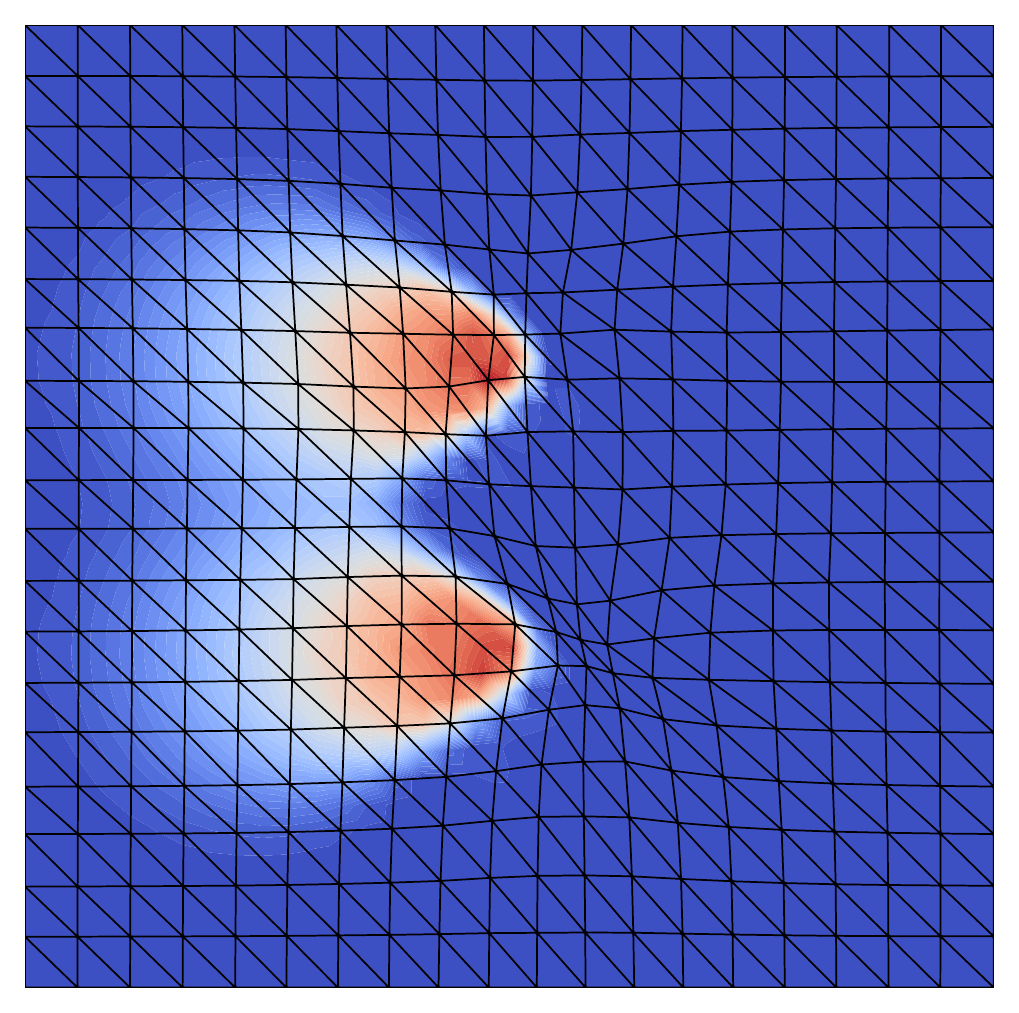}
    \end{subfigure}
\quad
    \begin{subfigure}{0.2\textwidth}
         \includegraphics[width=\textwidth]{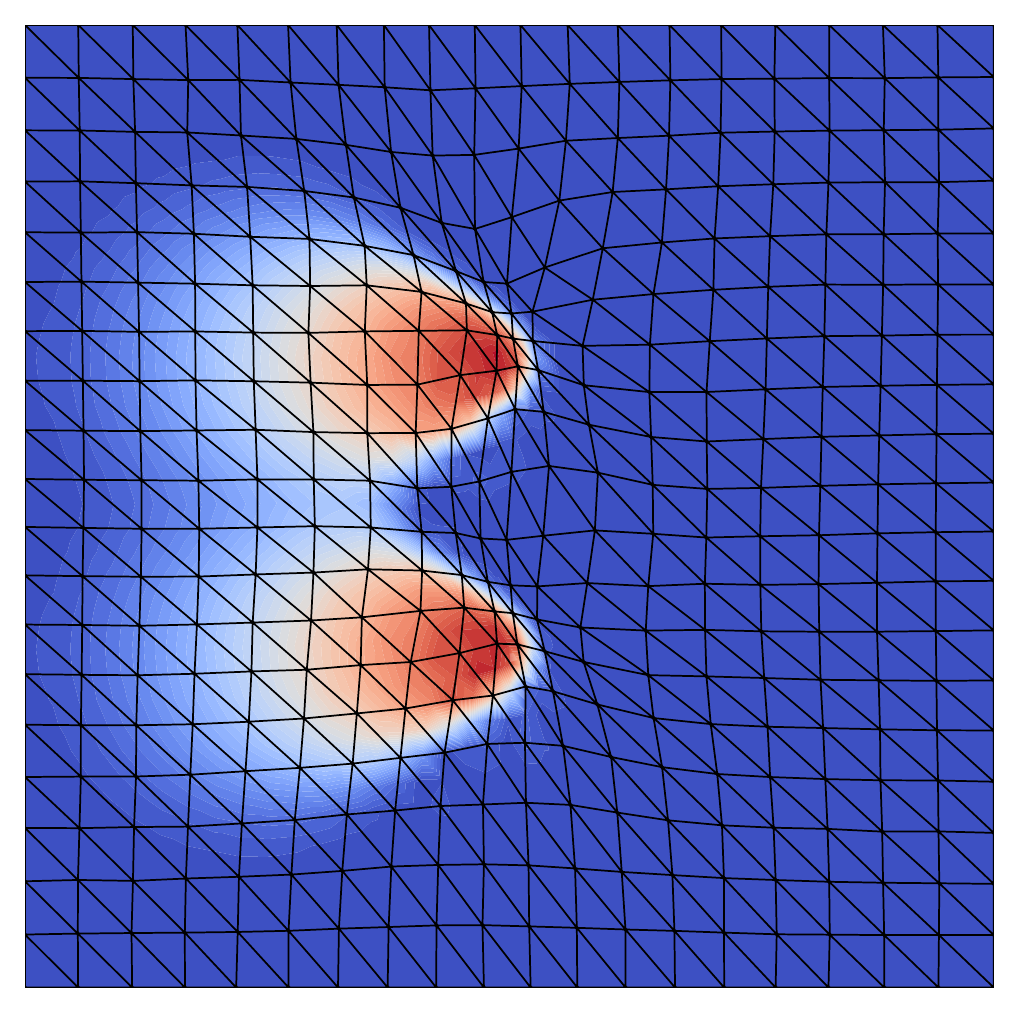}
    \end{subfigure}
\\
    \begin{subfigure}{0.2\textwidth}
        \includegraphics[width=\textwidth]{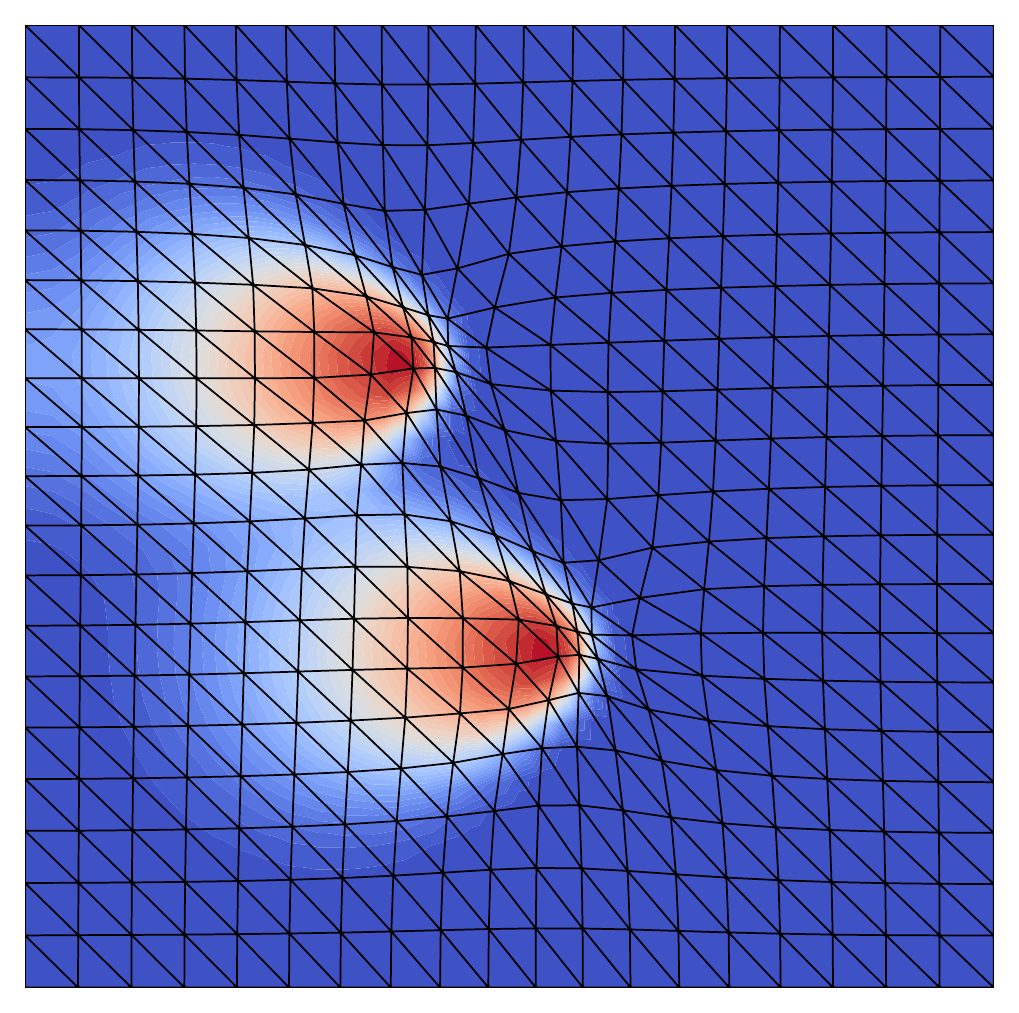}
    \end{subfigure}
\quad
    \begin{subfigure}{0.2\textwidth}
        \includegraphics[width=\textwidth]{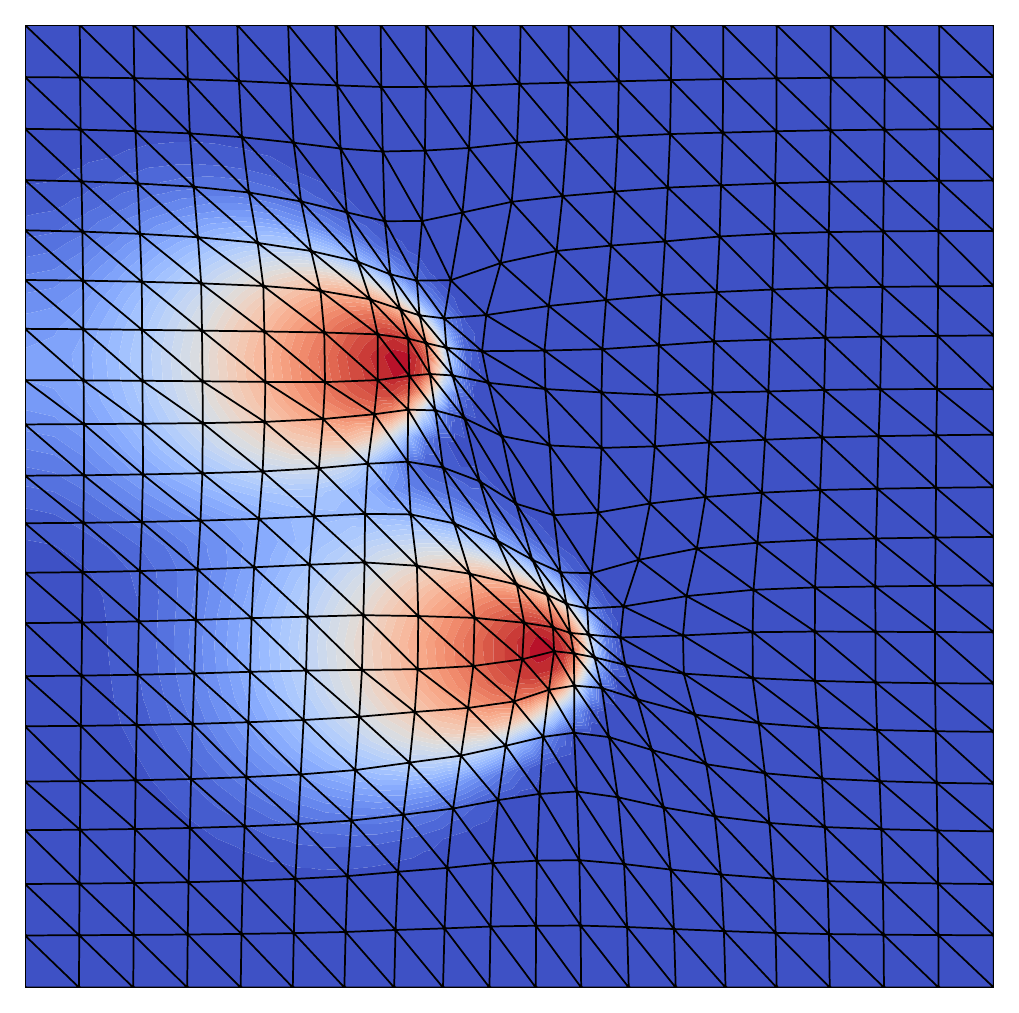}
    \end{subfigure}
\quad
    \begin{subfigure}{0.2\textwidth}
         \includegraphics[width=\textwidth]{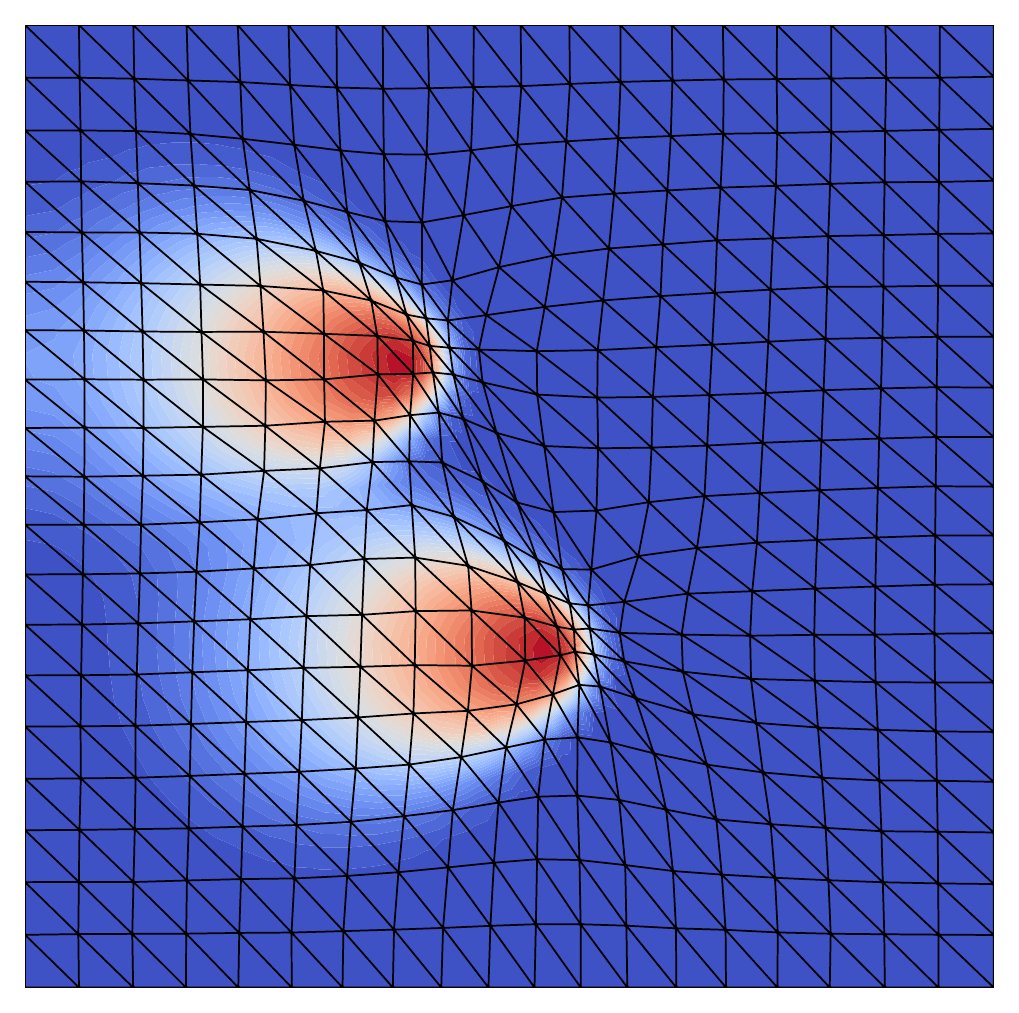}
    \end{subfigure}
\\
    \begin{subfigure}{0.2\textwidth}
        \includegraphics[width=\textwidth]{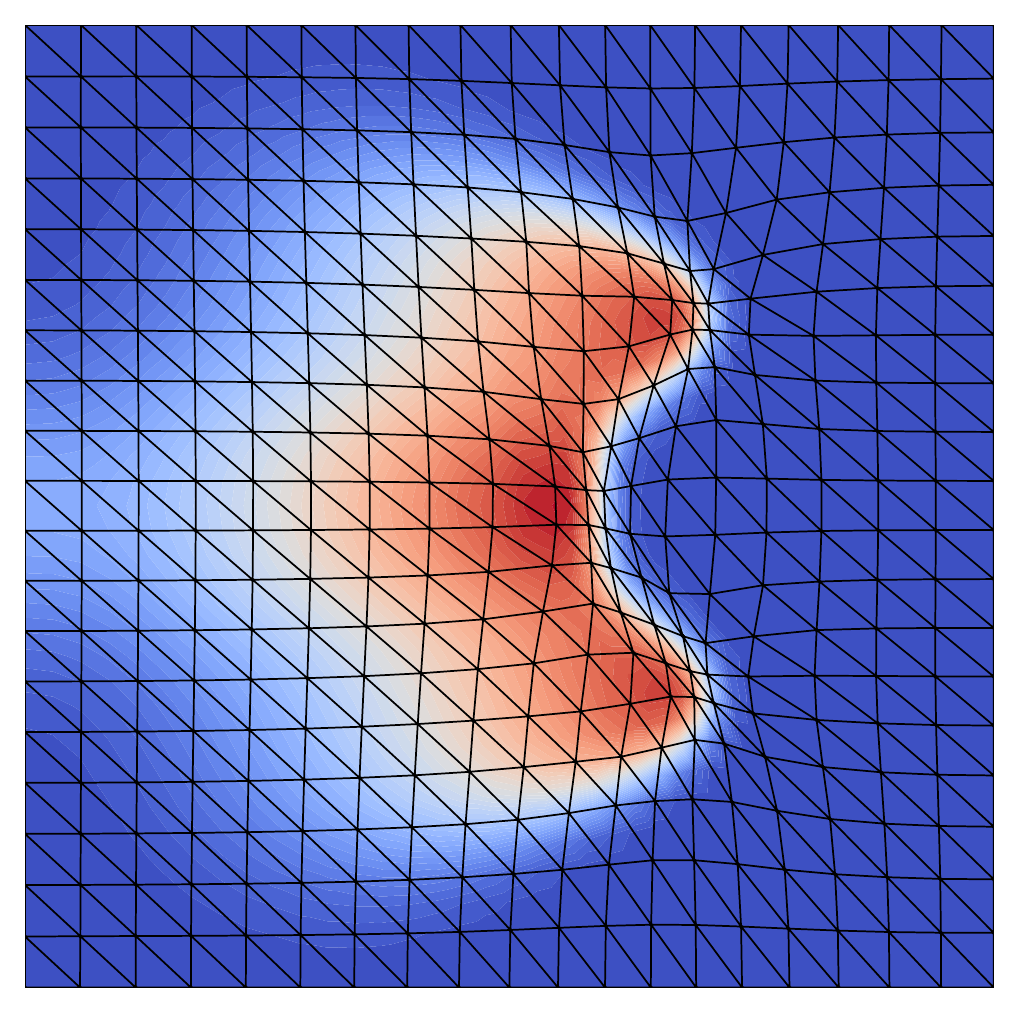}
        \caption{MA}
    \end{subfigure}
\quad
    \begin{subfigure}{0.2\textwidth}
        \includegraphics[width=\textwidth]{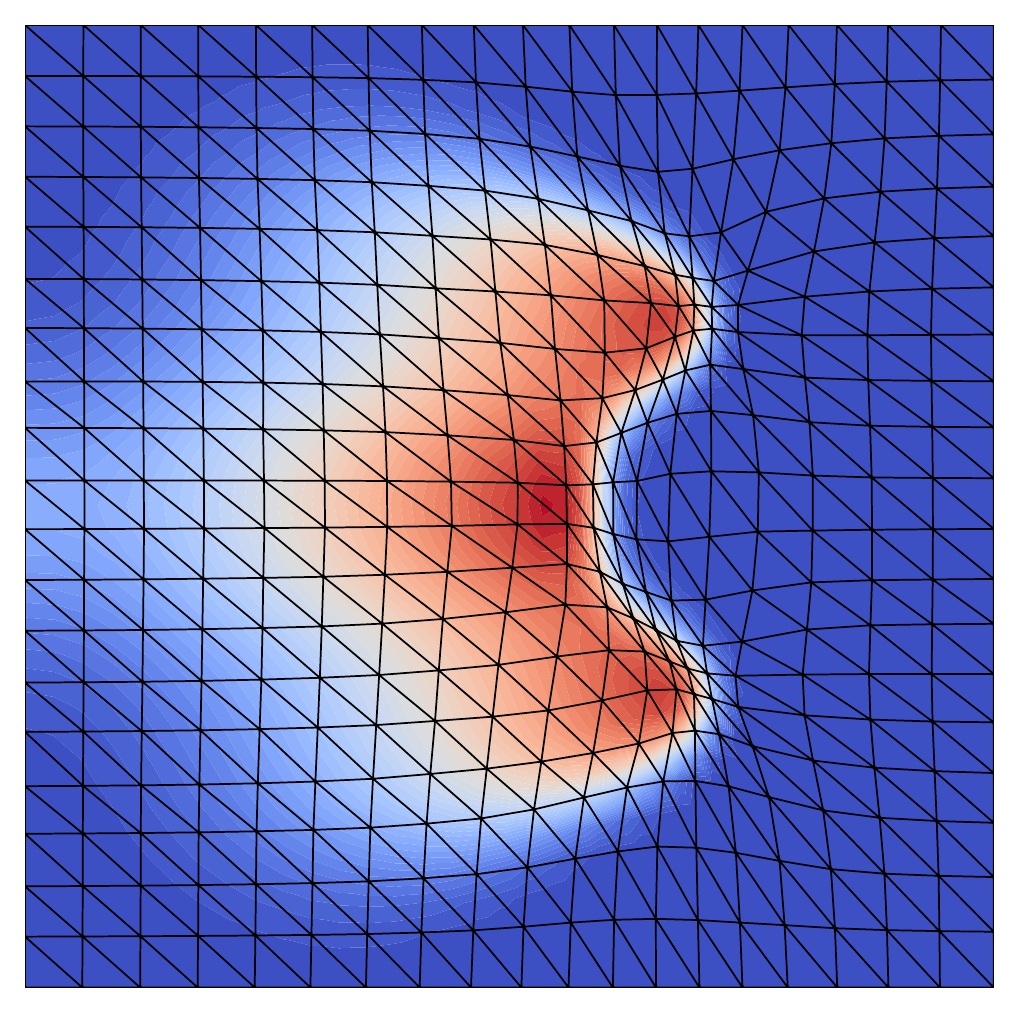}
        \caption{M2N-Spline}
    \end{subfigure}
\quad
    \begin{subfigure}{0.2\textwidth}
         \includegraphics[width=\textwidth]{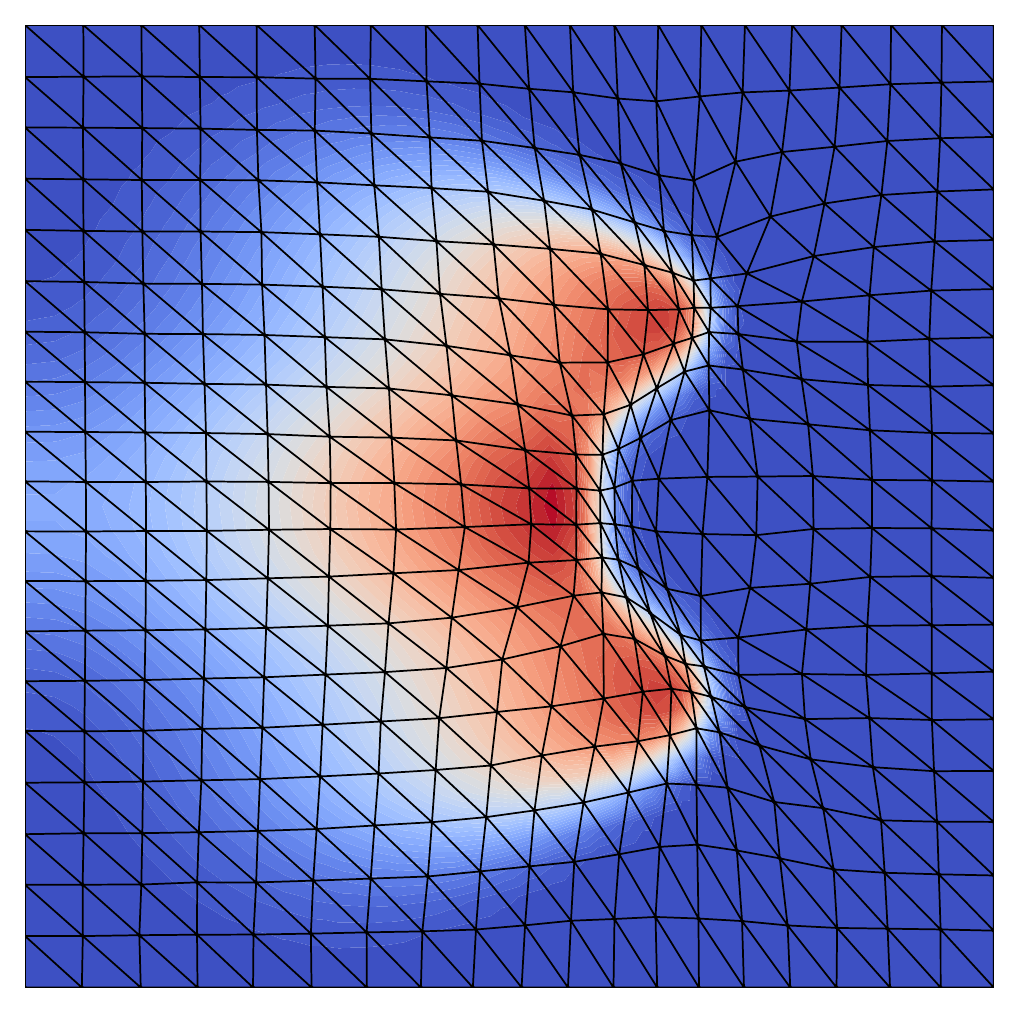}
         \caption{M2N-GAT}
    \end{subfigure}
\caption{Comparison of mesh movement for the Burgers' equation problem. In each row is a different sample.}
\label{fig:generalization_of_burgers}
\end{figure}

The results are summarized in Table \ref{tab:table_squared_burgers}. It can be found that all learning-based methods are three to four orders of magnitude faster than the traditional MA method. The acceleration is about one order of magnitude larger than the Poisson's equation experiments, because the MA method runs even slower for a nonlinear PDE. Mesh tangling never occurs for M2N-Spline and M2N-GAT in this experiment. The GNN-based models perform better than the other two models, because they contain extra local feature extractors, and local information can be propagated better through edges, which is important for the Burgers' equation problem. Some generated mesh examples at different timesteps of four different trajectories are shown in Figure \ref{fig:generalization_of_burgers}, where it can be seen that M2N-GAT is better at performing delicate local deformation compared to the M2N-Spline model.

\section{Conclusion}
In this paper we have proposed the Mesh Movement Network (M2N), which to the best of our knowledge is the first learning-based end-to-end mesh movement method for PDE solvers. Traditional mesh movement methods can improve the accuracy of numerical PDE solutions without modifying the topology of the mesh, at the expenses of solving an auxiliary PDE, which is often very computationally expensive and sometimes makes the approach infeasible. With the power of deep learning, M2N generates adapted meshes for different PDE problems of the same type, with the solution precision comparable to ground truth but at a much faster speed. To achieve this robustly, we have designed a Neural-Spline based and a GAT based mesh deformer, to guarantee the output adapted mesh retains boundary consistency, alleviates mesh tangling, and generalizes to different mesh densities. The results are validated on the static linear Poisson's equation with regular and irregular domains, and the time-dependent nonlinear Burgers' equation.

\bibliographystyle{unsrtnat}
\bibliography{references}

\end{document}